# Class-Imbalance-Aware Adaptive Dataset Distillation for Scalable Pretrained Model on Credit Scoring

Xia Li*[a,c,$$], Hanghang Zheng[b,$$], Xiwei Zhuang[a], Zhong Wang[a], Xiao Chen[a], Hong Liu[a], Jasmine Bai[c], Mao Mao[d]

[a] Risk Management Division, China Development Bank, 100031, Beijing, China
[b] School of Finance, Central University of Finance and Economics, 102206, Beijing, China
[c] GF Holdings (Hong Kong) Corporation Limited, Hongkong SAR, China
[d] University of Cambridge, Cambridge CB2 1TN, United Kingdom

[$$] Authors Contribute Equally
*Corresponding Author

## Abstract

The advent of artificial intelligence has significantly enhanced credit scoring technologies. Despite the remarkable efficacy of advanced deep learning models, mainstream adoption continues to favor tree-structured models due to their robust predictive performance on tabular data. Although pretrained models have seen considerable development, the use of such models for tabular-structured credit scoring datasets remains largely unexplored. Tabular-oriented large models, such as TabPFN, have made the application of large models in credit scoring feasible, albeit they can only process limited sample sizes. This paper proposes a novel framework that combines a tabular-tailored dataset distillation technique with a pretrained model, thereby improving the scalability of TabPFN. Furthermore, although class imbalance is a common characteristic of financial datasets, its influence during dataset distillation has not been systematically explored; we therefore integrate imbalance-aware objectives into the distillation process, resulting in improved performance on imbalanced credit data. Experiments on six publicly available credit scoring datasets show that the proposed imbalance-aware distillation improves AUC by up to 8.7 percentage points over the standard MSE-based distillation baseline. At the same time, distilled datasets retain roughly 76 – 95% of the full-data AUC while often using well under 10% of the original training samples (and at most 31.3%) of the original training samples and consistently outperforming random subsets. Taken together, these results demonstrate that imbalance-aware dataset distillation provides a compact yet accurate solution for imbalanced credit scoring and enhances the practical scalability of tabular pretrained models such as TabPFN.

# 1. Introduction

The field of financial technology has experienced significant advancements due to the emergence of deep learning and machine learning(Nazareth & Reddy, 2023). These computational paradigms have enabled data-driven decision-making across various financial sectors, supporting applications such as stock movement forecasting(Y. Li & Pan, 2022), credit risk evaluation(Dastile, Celik, & Potsane, 2020), and antimoney-laundering detection(Kute, Pradhan, Shukla, & Alamri, 2021). In contrast to stock forecasting, which primarily relies on time-series data, both credit scoring and money-laundering detection typically depend on multimodal information(Zhang & Song, 2022),(Cheng et al., 2021) represented in structured tabular form(Qian, Ma, Gao, & Song, 2023)·(Xu, Cirit, Asadi, Sun, & Wang, 2024). The domain of credit scoring has garnered considerable attention as customer data repositories have become more robust and comprehensive. As financial institutions accumulate increasingly comprehensive customer information, credit scoring—used as estimating a borrower's probability of default (PD) within a one-year horizon—has become a crucial component of risk management systems.

Credit scoring exhibits several distinguishing characteristics that pose practical and methodological challenges. **First**, its underlying data are typically tabular and highly imbalanced: default events are rare, often occurring at rates between 0.5%(Demajo, Vella, & Dingli, 2020),(Thomas, Crook, & Edelman, 2017) and 10%(Demajo et al., 2020),(Andreeva, Calabrese, & Osmetti, 2016) depending on sector, geography, and borrower type. Such rarity leads to significant class-imbalance problems, causing minority patterns to be easily overshadowed by majority trends(Dastile et al., 2020),(He & Garcia, 2009). **Second**, credit-risk data involve sensitive personal and financial information, giving rise to strong privacy-preservation requirements and strict regulatory constraints. Data sharing across institutions is generally infeasible, and high-quality labeled data are costly to obtain(Jing et al., 2021), further reinforcing privacy concerns. **Third,** credit-scoring distributions vary substantially across industries and regions—for example, mortgage portfolios typically exhibit extremely low default rates(Demajo et al., 2020),(Thomas et al., 2017), while SMEs in Europe may present much higher default risks(Demajo et al., 2020),(Andreeva et al., 2016). These domain discrepancies undermine the generalizability of models trained on any single financial application scenario.

Recent research has explored strategies to address these challenges individually. Techniques such as over-sampling(Gosain & Sardana, 2017), undersampling(Devi, Biswas, & Purkayastha, 2020) have been used to mitigate data level imbalance; synthetic data generation(Lu et al., 2023), dataset distillation(Yu, Liu, & Wang, 2023), and federated learning(Gong et al., 2022) have been applied to enhance privacy protection; and pretrained large models have been introduced to improve cross-domain generalization(Kim, Wang, Sclaroff, & Saenko, 2022). However, most pretrained models are designed for text or vision modalities, limiting their usefulness for tabular credit-risk tasks. The Tabular Prior-Data Fitted Network (TabPFN)—a transformer-based pretrained model tailored for tabular data—offers strong performance(Hollmann, Müller, Eggensperger, & Hutter, 2022), but its inference scalability is constrained by a strict upper limit on input size, making it difficult to apply directly in large-scale credit-scoring scenarios(Feuer, Hegde, & Cohen, 2023),(Magadán, Roldán-Gómez, Granda, & Suárez, 2023).

Taken together, these observations highlight three core challenges that are rarely addressed simultaneously in existing credit-scoring methodologies:(1) severe class imbalance, often tackled with re-sampling methods (2) strict privacy and data-sharing restrictions, motivating the need for synthetic or

distilled datasets but rarely incorporating imbalance handling; and (3) the need for scalable tabular priors, for which TabPFN is well suited, but whose deployment is limited by input-size constraints. **Integrating pretrained tabular models, ensuring privacy, and improving performance under heavy imbalance therefore remains an unresolved and practically important problem.**

To address these challenges jointly, we propose a class-imbalance-aware dataset distillation framework that integrates imbalance-handling principles directly into the distillation process. Our method augments Kernel Inducing Points (KIP) – based dataset distillation(Nguyen, Chen, & Lee, 2020) with class-balanced objective functions, enabling the distilled synthetic dataset to preserve minority-class geometry while providing only a small number of representative samples. Such synthetic datasets(Pan, Gong, Feng, & Li, 2024) are intended to support privacy-conscious data exchange: later NNDR analysis shows that distilled prototypes remain geometrically detached from individual records.

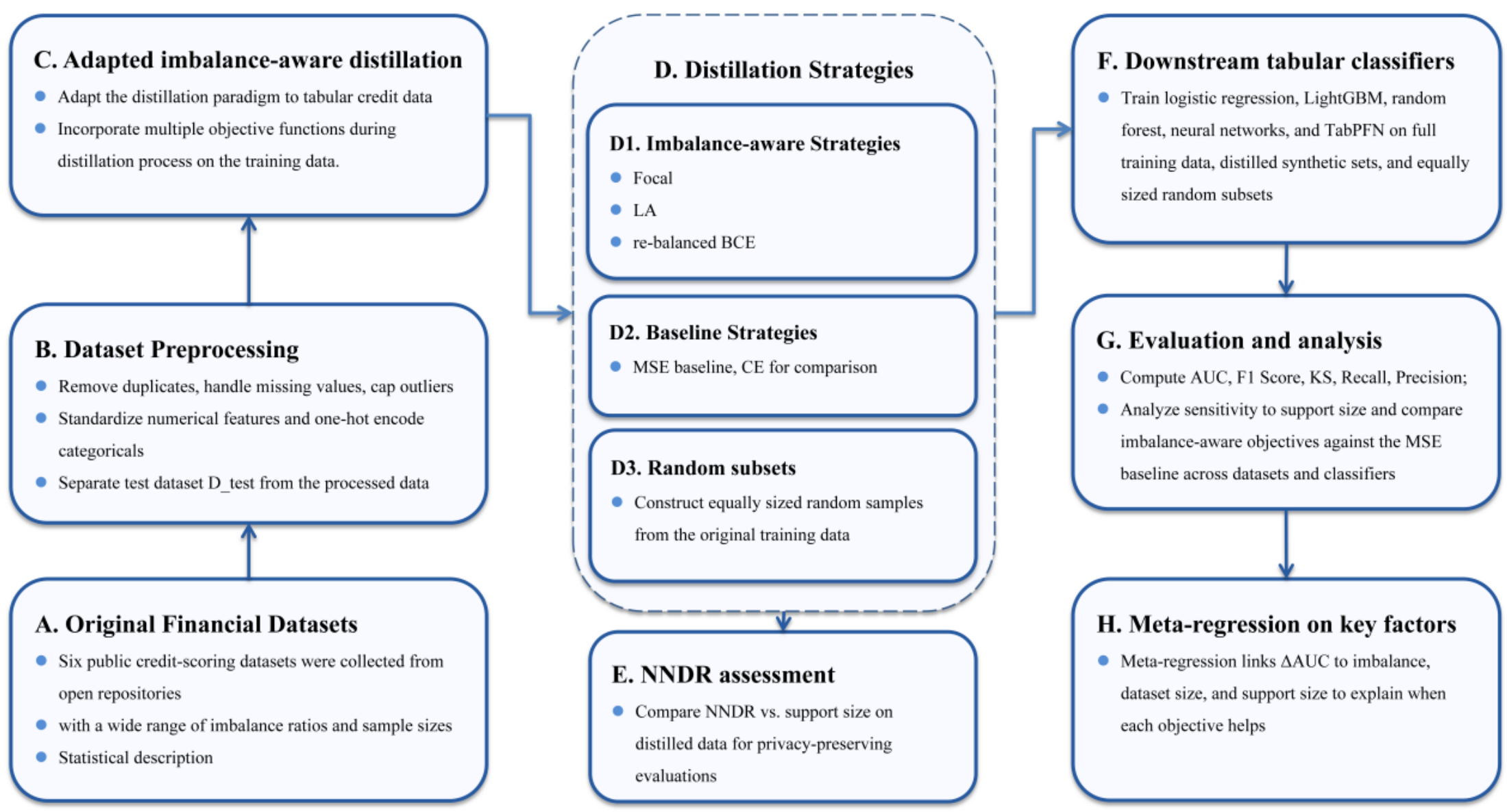

**Figure 1.** Overview of the proposed imbalance-aware dataset distillation framework for credit scoring. **A – B** show the curation and preprocessing of six public credit-scoring datasets, including construction of the test datasets. **C – D** depict the proposed imbalance-aware distillation stage, in which a small synthetic support set is optimized under several loss formulations. **E** assesses the privacy the distilled samples using NNDR across support sizes. **F – G** train and evaluate a diverse set of classifiers with discrimination metrics. **H** summarises a meta-regression analysis that links the performance gain over the MSE baseline.

**Figure 1** summarizes the overall logic of the study. Starting from six publicly available credit-scoring datasets with varying imbalance ratios, we construct a unified pipeline in which each dataset is split into training and held-out test sets (20% of the original samples). On the training portion, we learn a compact synthetic support set using the proposed imbalance-aware dataset distillation strategies, and then train a family of standard tabular classifiers either on the original data, on the distilled support sets, or on equally sized random subsets. We evaluate these models with AUC, KS, F1-Score, recall, and precision, and quantify the privacy property using NNDR. A meta-regression analysis was carried out on the change in AUC between imbalance-aware and MSE-based distillation to characterize the regimes.

In real-world credit scoring, practitioners must simultaneously cope with severe class imbalance, strict

privacy constraints, and large-scale tabular datasets that often exceed the limits of pretrained tabular models.

In summary, this work makes three main contributions to credit scoring with tabular pretrained models:

**1. A scalable, privacy-conscious framework for credit scoring.**
We introduce a tabular-tailored dataset distillation framework that compresses large, imbalanced credit datasets into compact synthetic supports, enabling pretrained tabular models such as TabPFN to be applied beyond their native sample-size limits. Across six public credit-scoring datasets, the distilled sets typically retain 76–95% of the full-data AUC while using only a small fraction of the original samples and consistently outperform random subsets.

**2. Imbalance-aware distillation objectives for better minority modeling.**
We integrate class-sensitive objectives directly into the distillation stage, rather than only at classifier training process. This imbalance-aware distillation yields consistent gains over an MSE-based baseline, improving AUC by up to 8.7 percentage points.

**3. Geometric privacy analysis and comprehensive empirical evaluation.**
We analyze the geometric relationship between distilled prototypes and original samples using the nearest-neighbour distance ratio (NNDR). Median NNDR values in the 0.8–0.9 range across objectives and support sizes indicate that distilled samples remain geometrically detached from individual records, providing geometric evidence of reduced memorization risk and supporting privacy-conscious data sharing, although formal differential privacy guarantees are left for future work. We further corroborate the effectiveness of the framework through extensive ablations on support size, loss functions, and baselines across multiple classifiers.

## 2. Literature review

Traditional credit scoring protocols, such as survival analysis(Bellotti & Crook, 2009) and statistical methods have long been the cornerstone for assessing creditworthiness. Machine learning methods, including LightGBM(LightGBM)(Ke et al., 2017),(Zhou, Fujita, Ding, & Ma, 2021), XGBoost(XGB)(Chen & Guestrin, 2016), Random Forest(Y. Wang, Zhang, Lu, & Yu, 2020),(Trivedi, 2020) along with logistic regression(H. Li & Wu, 2024) and their derivatives(Dumitrescu, Hué, Hurlin, & Tokpavi, 2022),(Moscato, Picariello, & Sperlí, 2021), have emerged as dominant players in credit scoring. Logistic Regression is preferred due to its good explainability, tree-based LightGBM/XGB is known for its superior performance for large scaled tabular dataset(Shwartz-Ziv & Armon, 2022). These methods enhance predictive power by adeptly capturing non-linear patterns and interactions among features. However, these models may degrade under small-sample or severely imbalanced credit-risk scenarios, motivating the exploration of pretraining techniques(Jing et al., 2021),(Dumitrescu et al., 2022) and data augmentation(Teng, Wang, Yang, Chen, & Li, 2023) to better leverage limited, noisy, or heterogeneous financial data.

The rapid advancement of pretrained large models has further broadened opportunities in financial analytics(H. Zhao et al., 2024). General-purpose models such as GPTs(Y. Cao & Zhai, 2023), BERT(Devlin, Chang, Lee, & Toutanova, 2018),(Hiew et al., 2019), RoBERTa(Y. Liu et al., 2019),(L.

Zhao, Li, Zheng, & Zhang, 2021), as well as finance-specific variants like FinGPT(Yang, Liu, & Wang, 2023) and BloombergGPT(Wu et al., 2023) have demonstrated their potential to revolutionize various facets of the fintech industry, including risk assessment(A. Huang, Qiu, & Li, 2021) and sentiment analysis(L. Zhao et al., 2021). Despite these successes, the deployment of pretrained large models in credit scoring remains limited, partly because most existing pretrained models are optimized for text or vision tasks rather than tabular learning. TabPFN(Hollmann et al., 2022)—a transformer-based model pretrained on synthetic tabular distributions—offers a promising direction by enabling one-shot inference and strong generalization. Nevertheless, its practical use is constrained by its processing data size(S.-Y. Liu & Ye, 2025),(Ruiz-Villafranca, Roldán-Gómez, Gómez, Carrillo-Mondéjar, & Martinez, 2024),(Hollmann et al., 2022), which limits scalability to real-world financial datasets with large sample sizes and many features.

To improve credit-scoring performance under class imbalance, recent research on state-of-the-art imbalance handling has focused on loss shaping and re-weighting strategies. Focal loss emphasizes hard minority samples by down-weighting easy majority instances(Lin, Goyal, Girshick, He, & Dollár, 2017); LDAM introduces label-distribution-aware margins that enlarge decision boundaries for minority classes under imbalance(K. Cao, Wei, Gaidon, Arechiga, & Ma, 2019); asymmetric and generalized extreme value-based algorithms explicitly encode class skew in the objective function and have been shown to improve model performance in imbalanced credit-scoring applications(X. Li, Zheng, Tao, & Mao, 2025),(Mushava & Murray, 2024); and class-balanced method adjusts weights based on the effective number of samples to avoid instability under skewed distribution(C. Wang, Deng, & Wang, 2020). However, these imbalance-handling techniques generally carry out after the training dataset has been fixed, and thus do not improve the informational structure of the training data itself. These loss-based and calibration-aware approaches represent the current state of the art in handling class imbalance and form the methodological basis on which our imbalance-aware dataset distillation is built.

Data-level methods such as Synthetic Minority Oversampling Technique (SMOTE)(Fernández, Garcia, Herrera, & Chawla, 2018) and random under-sampling (RUS)(Elhassan & Aljurf, 2016), together with their variants including ADASYN(He, Bai, Garcia, & Li, 2008), and hybrid schemes such as SMOTE-Tomek links, attempt to re-balance the dataset before training, yet they may inadvertently introduce noise, discard informative samples which lead to overfitting or underfitting(Almubark, 2024), or raise privacy concerns by relying on raw input features(H. Liu, Mao, Li, & Gao, 2025). Importantly, none of these methods explicitly address the challenges posed by privacy-sensitive, large-scale financial datasets, limited model scalability, and extreme class imbalance in a unified way.

Dataset distillation has emerged as a promising alternative for addressing data scalability, computational cost, and privacy constraints(Yu et al., 2023),(H. Liu et al., 2025) simultaneously. The goal of dataset distillation is to condense a large dataset into a much smaller synthetic set that preserves the essential training signal(Lei & Tao, 2023). This approach is particularly attractive for financial applications where datasets are large, costly to share, and sensitive in nature. Distillation techniques can reduce memory and computational requirements, facilitate knowledge transfer, and offer improved privacy protection(Pan et al., 2024) because only synthetic samples are shared. Existing distillation frameworks such as kernel ridge-regression(Nguyen et al., 2020), meta-learning-based methods such as MTT(B. Zhao, Mopuri, & Bilen, 2020), and gradient-matching have shown effectiveness on balanced datasets but remain

imbalance-agnostic, often reproducing majority-class patterns while neglecting subtle minority structures(W. Liu, Fan, Xia, & Xia, 2022).

Compared with prior work, our method introduces imbalance-awareness directly into the distillation stage. Instead of applying focal, asymmetric, or class-balanced losses only during classifier training, we incorporate them into the objective of distilling the synthetic dataset itself. As a result, the distilled set encodes minority-class geometry, enabling downstream learners — including logistic regression, LightGBM, MLP, KNN, and TabPFN—to better identify rare but critical default signals using far fewer samples. This design is particularly advantageous in privacy-constrained lending environments, where sharing raw data is infeasible but exchanging a small synthetic dataset is acceptable. Because financial datasets vary widely in scale and distributional characteristics, dataset distillation also provides a natural mechanism for generating compact representations that align with TabPFN's input-size limitations.

In summary, existing methods treat class imbalance, privacy, and tabular-model scalability as separate issues. Our work bridges these gaps by proposing a class-imbalance-aware dataset distillation framework that integrates focal, asymmetric, and class-balanced objectives directly into KIP optimization. This ensures that minority-class structure is preserved prior to model training, leading to improved downstream performance across diverse credit-scoring datasets.

**Comparison with existing work**

Existing dataset distillation methods have been mainly developed for image domains(Kang, Ram, Zhou, Samulowitz, & Seneviratne, 2024) and not specifically designed for highly imbalanced tabular credit data, and do not incorporate imbalance-aware objectives during the distillation stage. In parallel, the class-imbalance literature mostly focuses on modifying the classifier loss (e.g., focal or class-balanced loss) or mixed-resampling strategies on the original data(X. Huang, Zhang, & Yuan, 2020),(Alamri & Ykhlef, 2024), rather than on generating synthetic condensed datasets. The credit scoring research either adopt ensemble tree-based models(Ali, Khedr, El-Bannany, & Kanakkayil, 2023) and few research combine dataset distillation with imbalance-aware objectives in processing tabular credit scoring tasks. In contrast, our work explicitly integrates imbalance-aware losses into the distillation process for financial credit datasets, and systematically evaluates their effect across multiple classifiers, including tree-based models and TabPFN.

## 3. Experiments

In this section, we evaluate the proposed imbalance-aware dataset distillation framework on several credit-scoring datasets. We follow the workflow summarized in **Fig 1**. We first describe the datasets and preprocessing procedures, and then present the experimental setup and results. Details are shown as follows.

### 3.1 Dataset preprocessing

The datasets selected for this study encompass a variety of financial data, sourced from open platform Kaggle (listed in **Table 1**). The imbalance ratio (***IR***) —defined as the quotient of negative (non-default) to positive (default) instances—fluctuates between ~1.7 to ~14.9, representing the biased distribution patterns of financial industry. The datasets are available on the Kaggle platform and the details of statistics of the datasets are listed in **Table 1.**

| Dataset | Samples | Defaulters | Features | ***IR*** |
|---|---|---|---|---|
| GMC | 150000 | 10026 | 11 | 14.9 |
| Credit Card | 30000 | 6630 | 24 | 3.5 |
| CHN Banks | 60780 | 10586 | 38 | 5.7 |
| Lending Club | 14785 | 5475 | 54 | 1.7 |
| Bank Personal Loan | 3200 | 307 | 13 | 9.4 |
| German Credit | 640 | 192 | 21 | 2.3 |

**Table 1.** Statistical description of various datasets and their profiles including IRs.

The GMC dataset contains 150,000 consumer credit profiles with 11 numerical financial features (e.g., debt ratio, income, delinquency counts) and exhibits a high imbalance ratio (IR~14.9). The Credit Card dataset includes 30,000 clients with 24 numerical and categorical features describing billing and demographic information (IR~3.5). The CHN Banks dataset comprises 60,780 banking customers with 38 mixed demographic, transactional, and loan-related features (IR~5.7). The Lending Club dataset consists of 14,785 peer-to-peer loan cases with 55 heterogeneous borrower and loan attributes and the lowest imbalance severity (IR~1.7) but the highest feature dimensionality. The Bank Personal Loan dataset contains 3,200 customers with 13 demographic and account-level variables (e.g., income, family status, existing banking relationships) and exhibits a relatively high imbalance (IR~9.4). The German Credit dataset includes 640 credit applicants with 21 socio-economic and contract-related attributes and a moderate imbalance (IR~2.3). Taken together, these datasets span a wide range of sample sizes, feature spaces, and imbalance levels across consumer, banking, and peer-to-peer lending scenarios, providing a heterogeneous and realistic test dataset pool for evaluating imbalance-aware dataset distillation. A more detailed description can be seen in **Table S1.**

### 3.2 Baseline distillation objective and downstream models

Our primary baseline is the standard imbalance-agnostic MSE distillation objective used in original dataset distillation work. On top of this baseline, we consider several imbalance-aware distillation objectives, including focal loss, LA, and class-balanced binary cross-entropy variants, which explicitly encode class skew in the objective. All objectives share the same architecture, optimization procedure, support size, and datasets so that performance differences can be attributed to the choice of distillation objective.

Concretely, we start from the standard KIP objective using an MSE loss and treat this as our imbalance-agnostic baseline. Our imbalance-aware variants are obtained by replacing this MSE loss with focal, LA, class-balanced binary cross-entropy, while keeping the distilled support-set size, optimizer, and number of training steps identical. This makes the distillation process directly sensitive to class frequencies without changing the overall architecture or training pipeline.

For evaluation, we train standard credit-scoring classifiers on (1) the full training data, (2) distilled synthetic sets, and (3) equally sized random subsets. We use logistic regression, LightGBM, k-nearest neighbours (kNN), a shallow multilayer perceptron (MLP), and TabPFN as diverse downstream classifiers, relying on widely used public implementations, consistent preprocessing, and recommended hyperparameter settings. This setup isolates the effect of the distillation objective while providing a fair and reproducible comparison across modeling paradigms.

### 3.3 Imbalance-aware dataset distillation

Given that TabPFN typically exhibits optimal performance on small-scale datasets, we designated the sample size below 1000 for the output of the distillation. To construct compact training sets suitable for TabPFN, we employ a KIP-style dataset distillation framework. The goal is to optimize a synthetic dataset $\boldsymbol{D_s}$ = (Xs, ys) , |Ds| = m ≤ 1000, so that a lightweight proxy model trained on $\boldsymbol{D_s}$ reproduces the behavior of a model trained on the full dataset $\boldsymbol{D}$. In our experiments, we vary the size of this distilled training set—hereafter referred to as the support size, i.e., the number *m* of synthetic distilled samples—between 10 and 1000. Formally, the distillation objective is expressed as:

$$\min_{D_s} E_{(x,y)\sim D_{\mathrm{val}}} [\mathscr{L}(f_{D_s}(x),y)] \tag{1}$$

Where $f_{D_s}$ is the NTK-based kernel ridge regressor (KRR) trained on the synthetic set. This formulation encourages the synthetic set to contain the functional characteristics of the original data while remaining extremely small. Such that a kernel model trained on $\boldsymbol{D_s}$ approximates the behavior of a model trained on the full dataset.

Because credit-risk datasets exhibit strong class imbalance, we evaluate five loss functions within the above equation. In our experiments, We use the standard Mean Squared Error (MSE) objective strategy as our distillation baseline, which corresponds to the original dataset distillation formulation in as follow:

$$\mathscr{L}_{MSE}=\frac{1}{N}\sum_{i=1}^{N}(f_{D_i}(x_i)-y_i)^2 \quad （2）$$

It is noted that, under severe class imbalance, MSE treats all samples equally and therefore often biases the synthetic set toward majority-class behavior. This makes MSE a strong baseline but insufficient for minority-sensitive distillation.

**For classification problem, cross-entropy** loss (CE) is commonly used, CE is articulated as follows:

$$\mathscr{L}_{CE}=-\frac{1}{N}\sum_{j=1}^{N}[y_j\log p_j+(1-y_j)\log(1-p_j)] \tag{3}$$

where N denotes the number of samples, $y_j$ is the true binary label of sample j and $p_j$ is the predicted default probability for sample j obtained via the sigmoid function. In the credit-scoring context, pj can be interpreted as the model's estimate of how likely a given client is to default. The CE loss is minimized when these predicted probabilities are well aligned with the true labels.

The re-balanced CE variant adjusts for class imbalance by reweighting positive and negative samples according to their relative frequencies, **Re-weighted Balanced Cross Entropy** (re−Bn) is shown as follows:

$$w_+=\frac{1}{\pi}, \qquad w_-=\frac{1}{1-\pi} \tag{4}$$

$$\mathscr{L}_{re-Bn}=-\frac{1}{N}\sum_{i=1}^{N}[w_+y_i\log p_i+w_-(1-y_i)\log(1-p_i)] \tag{5}$$

Where $\boldsymbol{w}$ is the ratio of number of good client over number of defaulters of the original training set.

To emphasize hard examples, the **Focal Loss** is engineered to mitigate the challenges posed by class imbalance by intensifying the learning emphasis on difficult, misclassified instances(Lin et al., 2017).

For each sample, we compute

$$\mathscr{L}_{Focal}(y, p) = -\alpha(1 - p_t)^{\gamma} \log p_t \tag{6}$$

with fixed $\gamma = 2.0$ and $\alpha = 0.25$.

we also applied a logit adjustment classification algorithm (LA), which it combines the sigmoid adjusted factors with focal loss. Using the imbalance ratio (IR), it shifts the logits before applying the sigmoid

$$p_{adj} = \frac{1}{1+\exp(-Z+\boldsymbol{G}(\mathrm{IR},\alpha,\beta))} \tag{7}$$

$$G(IR, \alpha, \beta) = \alpha \mathrm{Ln}(\mathrm{IR}) + \beta \tag{8}$$

With α and β are hyper-parameters and IR is the Imbalance Ratio of the credit scoring dataset. The adjusted probability is computed and we re-use the focal loss function:

$$\mathscr{L}_{LA}(y, p_{adj}) = -\alpha(1 - p_t)^{\gamma} \log p_{\mathrm{j}} \tag{9}$$

All classification-based objectives (CE, re-Bn, focal, and LA) are plugged into the same distillation inner loop. Given the current distilled support set, we perform kernel regression, compute the chosen loss on the training batch, and through the kernel regression to update the support points. For each strategy(different objective functions) we run a small grid search over the learning rate (5e-4~1e-2), together with an patient criterion (min improvement ⩾1e-4) based on the validation loss and a epoch of 500, to select the best learning rate and effective training horizon for each dataset. The distillation was carried out on NVIDIA Tesla P100 graphics card with a memory of 16 GB.

### 3.4 Model training

Various models, including linear, tree-based and the pretrained classifiers TabPFN, were selected for the learning process on the distilled datasets. We first create an 80/20 train – test split for each dataset and perform 5-fold cross-validation on the training portion. For all classical models, we run lightweight grid searches around widely recommended default hyperparameters, whereas TabPFN is used in its standard inference configuration and only requires batching adjustments to satisfy input-size constraints. Random states are fixed to 42, and learning-rate candidates are explored in the range [0.01, 0.1] when applicable. Logistic Regression is configured with an L2 penalty, the liblinear solver, and 200 maximum iterations. The pretrained TabPFN classifier runs with batched inference (batch size 100), making it particularly well suited for small distilled sets (<1000 samples). Among classical tree and neural baselines, we include a Random Forest with 100 trees and unrestricted depth, LightGBM with 100 estimators and 31 leaves, and an MLP classifier with a single 100-unit hidden layer, ReLU activation, and Adam optimization for 200 iterations. The same data splits and training protocol are used for models trained on full data, distilled sets, and random subsets to ensure a fair comparison. All learning processes were run on environments with configurations of 25 vCPU Intel(R) Xeon(R) Platinum, NVIDIA Tesla P100 graphics card with a memory of 16 GB and PyTorch 2.8.0, Python 3.12.

We take model performance on the datasets distilled from CE as an example for an intuitive visualization, which is demonstrated in **Figure 2**. It can be seen the distilled dataset, only with 1% of the original dataset size, performed an approached AUC to the original dataset.

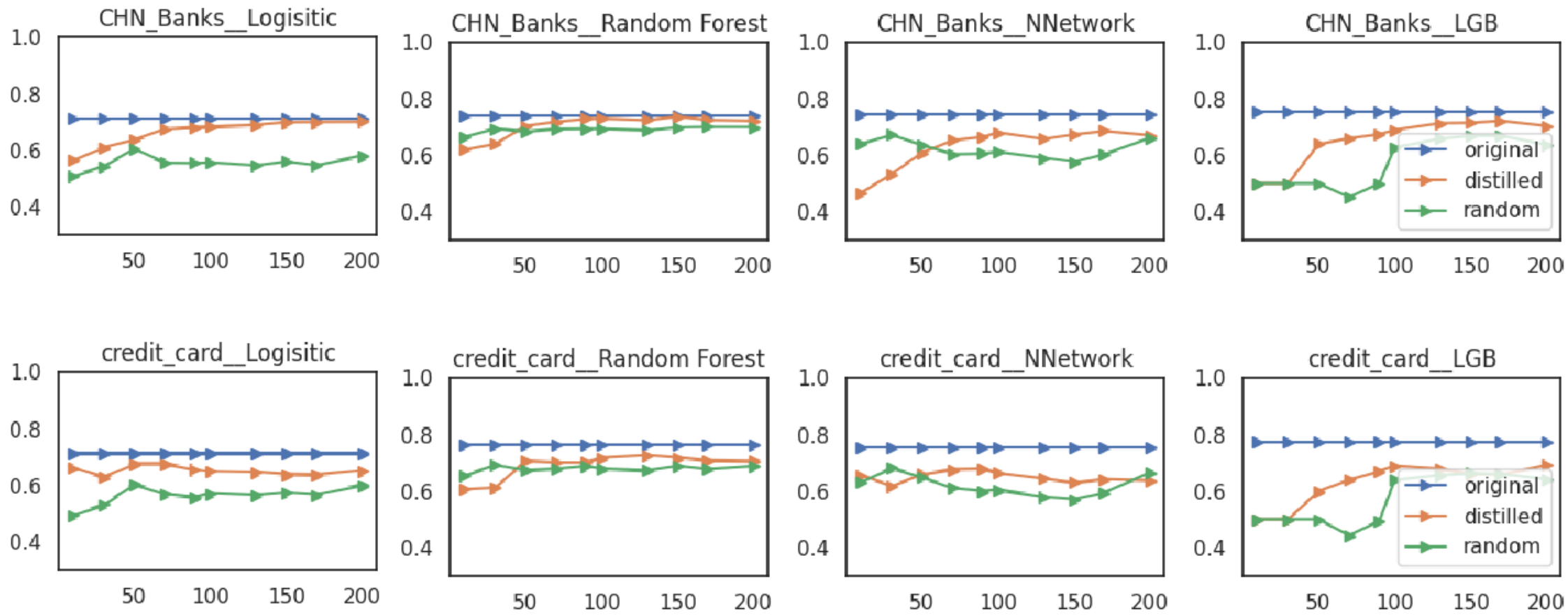

**Figure 2.** A demonstration of models performances in AUC on CHN Banks and the GMC datasets, where the distilled dataset was prepared using CE as objective function. The blue lines are the model performance on the original full dataset (with constant AUC).

### 3.5 Evaluation metrics and statistical testing

We select AUC (Area Under the ROC Curve) as the primary metric because it (1) is threshold-free, (2) remains relatively stable under class imbalance, and (3) reflects ranking quality that is directly relevant to credit cutoffs. However, AUC alone may obscure how well the model treats the minority (default) class; therefore, we complement it with the Kolmogorov–Smirnov (KS) statistic, recall, precision, and F1-score. AUC and KS primarily assess the separation and ranking between defaulters and non-defaulters, while recall, precision, and F1-score focus on the correctness of default predictions under imbalanced prevalence.

In addition, we also quantify how geometrically distinct the distilled samples are from the original data using the Nearest Neighbour Distance Ratio (NNDR). For each distilled sample, we compute the distance to its closest and second-closest neighbours in the original training set and take their ratio; higher NNDR values indicate that the distilled point does not lie extremely close to any single original record. We summarize the distribution of NNDR values across support sizes and objectives using boxplots in the main text and provide full results in the **Supporting Material**.

## 4. Discussion

We evaluate the proposed imbalance-aware distillation framework by comparing classifiers trained on full data, randomly subsampled data, and distilled datasets across heterogeneous credit-scoring tasks. Distilled subsets recover most of the predictive performance of full-data training and consistently outperform random subsets of the same size, indicating that they preserve decision-relevant structure under strong compression. Imbalance-aware objectives such as Focal and LA tend to improve over the MSE baseline, especially at small support sizes where data reduction is most severe. At the same time, NNDR analysis shows that distilled samples remain geometrically distinct from individual records, suggesting that the method abstracts rather than memorizes the raw data; the following subsections elaborate on these aspects in more detail.

### 4.1 Dataset distillation overview

We first demonstrate how different distillation strategies reshape the geometry of the synthetic datasets.

**Figure 3** traces the evolution of a distilled dataset As the distilled set size (support size) increases from 10 to 1000 samples. With only 10 – 50 distilled points, the PCA plots show that class clusters are sparse and many regions remain uncovered, whereas increasing the support size quickly fills in the structure of the original manifold and stabilizes the global layout well before 1000 samples.

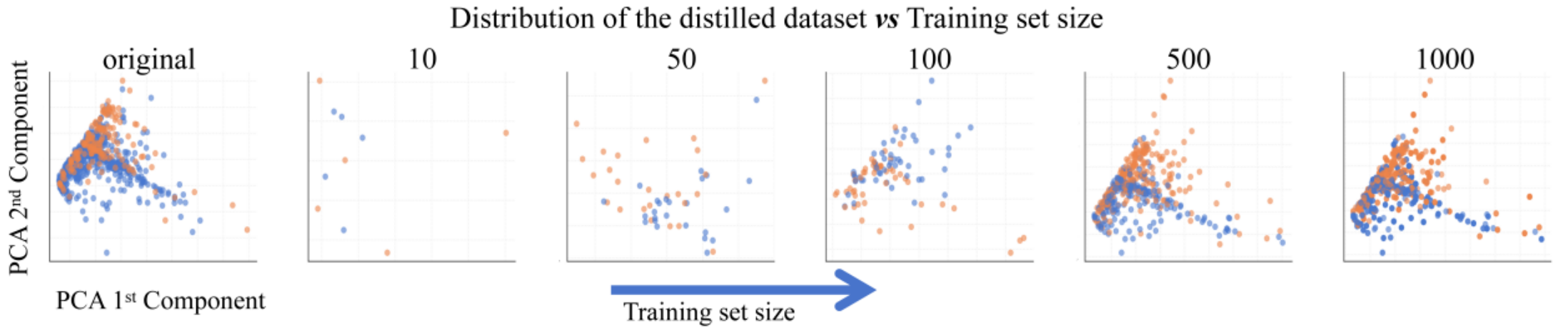

**Figure 3.** Evolution of a distilled dataset as the support size increases from 10 to 1000 samples. Each panel shows the PCA projection of the synthetic set for a given support size.

**Figure 4** extends this comparison across multiple datasets and objectives. The leftmost column shows the original training data, while the remaining columns display distilled datasets obtained with MSE, CE, Focal, LA, and re-Bn for the same support size. Across various sample datasets, all strategies recover the main class clusters and relative positioning of minority and majority classes; their PCA projections are visually very similar to the originals.

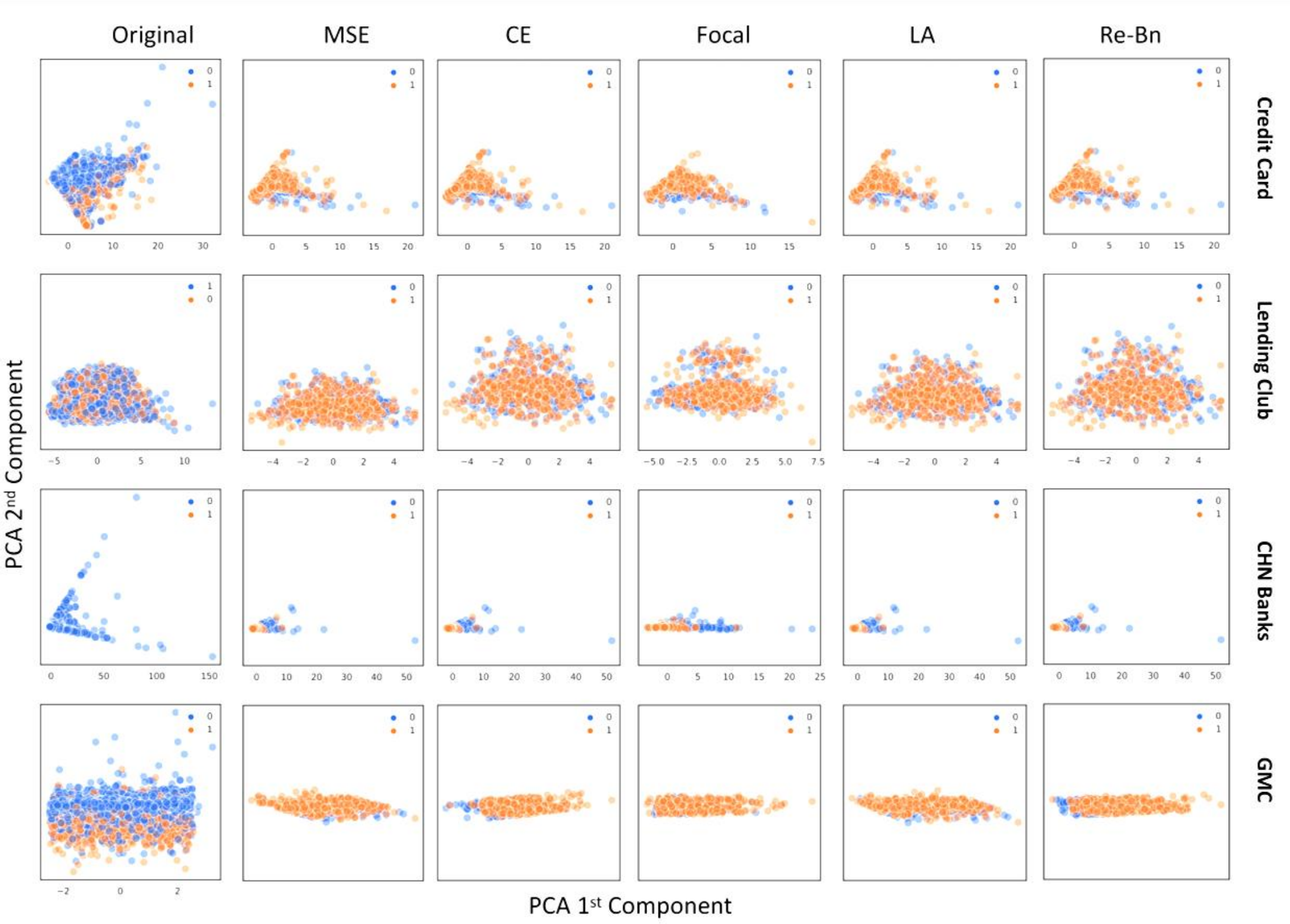

**Figure 4.** PCA projections of the original training data and distilled datasets obtained with different objective functions across multiple credit datasets.

This visual similarity, however, does not imply that the synthetic samples are trivial copies of individual records. To quantify how distinct distilled points remain from the raw data, we compute the NNDR between each distilled sample and its closest neighbour in the original dataset. Across all datasets and support sizes, the distilled samples consistently remain at a substantial geometric distance from

individual records, even though the two sets are visually similar in PCA space. For small to moderate support sizes (10 – 1000), the median NNDR across CE, Focal, LA, MSE and re-Bn is stably in the 0.8 – 0.9 range. Overall, these median NNDR values in the 0.8 – 0.9 range across objectives and support sizes indicate that distilled samples remain geometrically detached from individual records, providing geometric evidence of reduced memorization risk and suggesting improved privacy characteristics compared with direct sample sharing.

In contrast, SMOTE exhibits systematically lower NNDR because its augmented dataset consists of the original records themselves (for which NNDR necessarily collapses to zero) together with linearly interpolated points that lie very close to their source samples in feature space, and therefore provide virtually no geometric abstraction. The Random baseline is an even more extreme non-abstractive case, as it is simply a subsample of the raw data and all NNDR values degenerate to zero. The distilled datasets sit between these two extremes: they maintain strong task performance while keeping synthetic points well away from any individual record, achieving a more robust trade-off between privacy protection and statistical alignment with the original data. The NNDR evaluation of all six datasets can be found in Supporting Material.

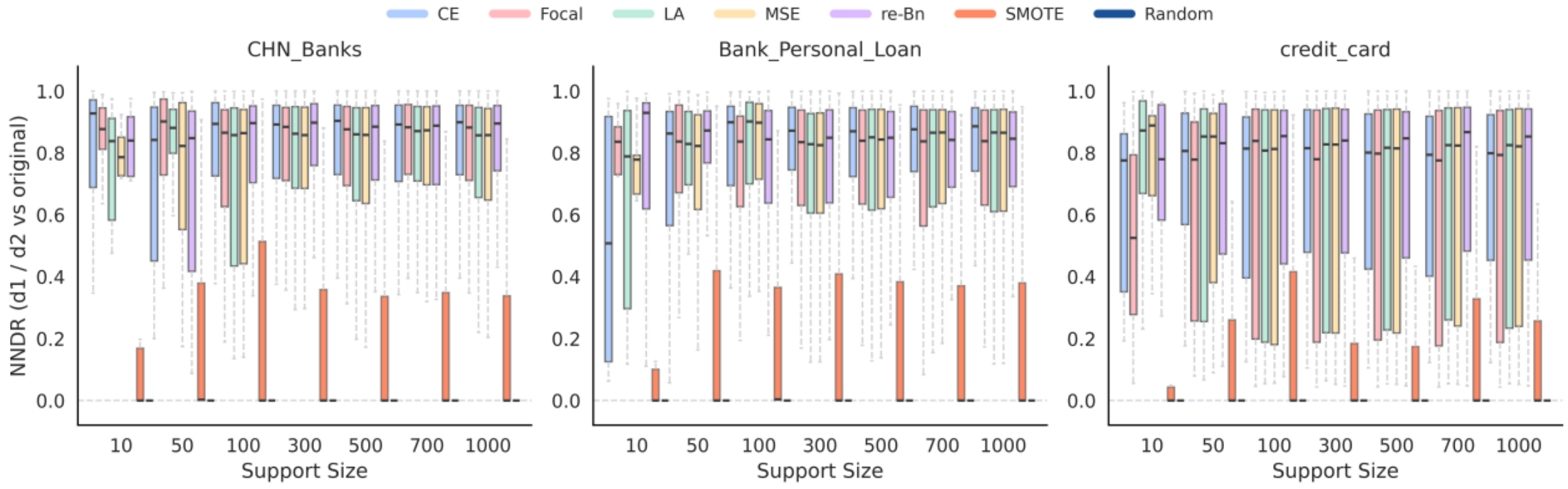

**Figure 5**. NNDR of distilled datasets across support sizes and distillation strategies. Each subplot displays NNDR distributions for a given dataset, computed between each distilled point and its closest original sample. Higher values indicate greater geometric distinctness from individual records. For clarity, the main figure includes three representative datasets; full results for all losses and support sizes are provided in the Supporting Material **Figure S1**.

### 4.2 Distillation computational efficiency

We then assessed the computational efficiency of the proposed distillation framework. Across all support sizes and datasets, the empirical average distillation runtimes were 2.08 ms for MSE, 2.09 ms for CE, 2.14 ms for re-Bn, 2.21 ms for Focal, and 2.11 ms for LA per distilled batch. MSE and CE achieve the fastest runtime, and the imbalance-aware objectives re-Bn, Focal, and LA introduce only modest overhead, increasing runtime by approximately 0.96 – 5.74% relative to CE due to class-dependent weighting and asymmetric activation terms. Importantly, even the slowest objective remains computationally lightweight, with all runtimes confined to the 2 – 2.3 ms range. Given that the distilled dataset size does not exceed 1000 samples, the overall distillation cost is negligible compared with training on full-scale datasets. These results confirm that the additional computational burden introduced by imbalance-aware objectives is minimal relative to the substantial performance gains they bring.

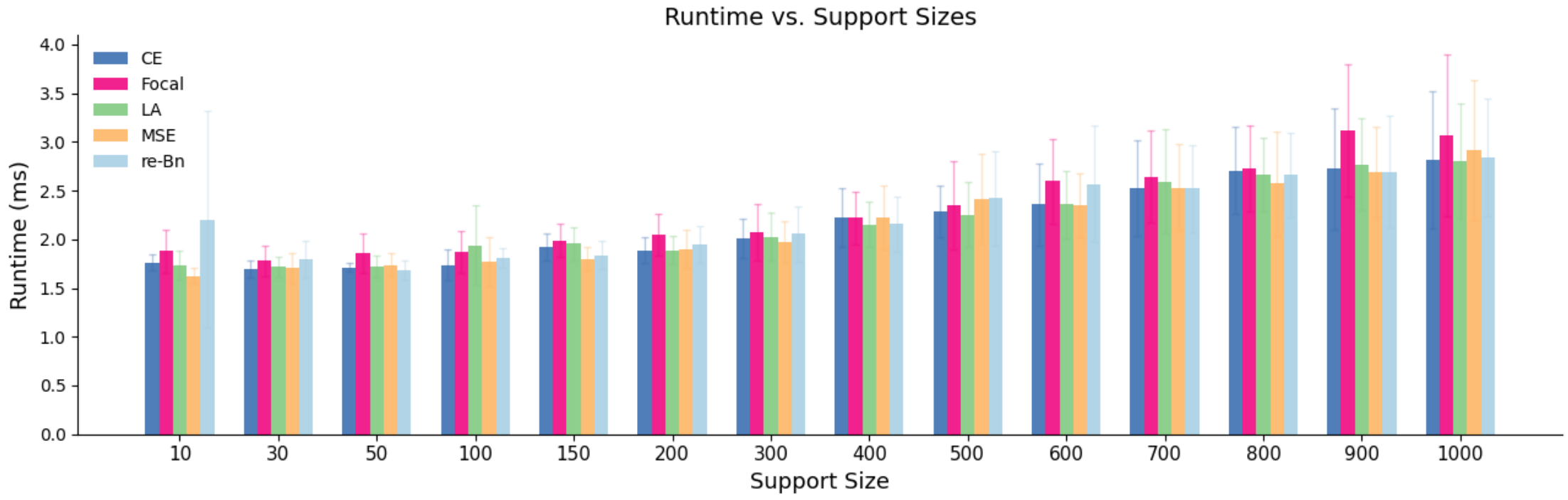

**Figure 6** Distillation runtime (ms) versus support size for five objective functions. Runtime remains low for all strategies, with MSE and CE being the fastest and Focal Loss slightly more expensive due to the extra weighting of minority samples. Details are shown in **Table S2.**

### 4.3 Model performances

To directly evaluate the impact of distilled sample size on model performance, we trained all classifiers on synthetic datasets with support sizes spanning 10 to 1000 samples. AUC rises sharply as the distilled sample size increases from 10 to approximately 100-300 samples; beyond this threshold, performance gains saturate, and curves converge toward the full-data baseline. **Table S3** and **Table S4** showed the distilled datasets retain roughly 76 – 95% of the full-data AUC while often using well under 10% of the original training samples (and at most 31.3%) of the original training samples and consistently outperforming random subsets.

#### 4.3.1 Distillation improves information content of compact datasets

Across all credit-scoring datasets, distilled subsets consistently narrow the performance gap between random sampling and full-data training. **Figure 7** summarizes model performance—averaged over all support sizes and classifiers (excluding TabPFN as there is no TabPFN trained on original full sized dataset)—for three training schemes: original data, distilled subsets, and random subsets. For example, on Bank_Personal_Loan, distilled subsets achieve an average AUC of ~0.92 (nearly matching the full-data model), while random subsets of the same size maintain an AUC of only ~0.65, corresponding to a substantial Δ AUC gain of ~0.28. Similar trends hold for the KS statistic: distilled subsets outperform random ones by up to 0.46 on Bank_Personal_Loan and by 0.05 – 0.12 on other datasets.

These gains are both quantitatively significant and configurationally consistent. Paired t-tests confirm that per-configuration differences in AUC and KS are statistically significant ($p < 10^{-3}$ for all datasets, often $p < 10^{-10}$). This indicates distilled samples retain the original training distribution's core decision boundary, whereas random subsampling fails to preserve this structure under moderate-to-severe class imbalance. Narrow error bars further verify that distilled data's superiority is robust to classifier choice and support size, confirming dataset distillation as an effective strategy for compressing large tabular credit-risk datasets. Notably, this compression preserves discriminative performance of its original datasets.

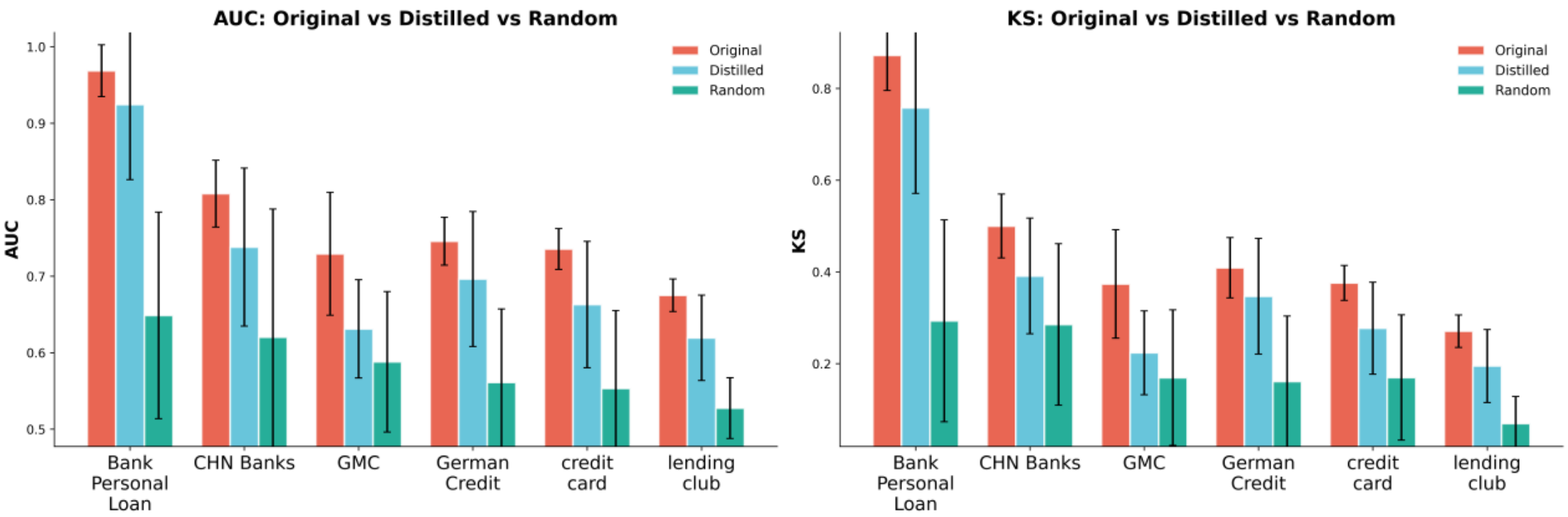

**Figure 7** Overview of performance for models trained on Original, Distilled, and Random datasets across all credit datasets, evaluated using AUC (left) and KS (right). Distilled datasets generally outperform randomly selected subsets and often approach the performance of models trained on the full original data, validating the effectiveness of the distillation strategy.

Table 2 shows the corresponding dataset-wise statistics, including $\Delta AUC_{random}$, $\Delta KS_{random}$, $\Delta$ F1-score$_{random}$, $\Delta Recall_{random}$, and $\Delta Precision_{random}$. Across all six credit datasets, distillation significantly outperforms random sampling in AUC, with $\Delta AUC_{random}$ ranging from +0.043 (GMC) to +0.276 (Bank Personal Loan), and all improvements being highly significant ($p < 0.001$, often far smaller, as indicated by ***). The gains are even more pronounced in F1-score and Recall, where distilled subsets improve minority detection by approximately 0.15 – 0.63 points in most datasets. Precision is also improved in nearly all cases, with the only mild exception being GMC, where a small drop in precision reflects the usual recall – precision trade-off under extreme imbalance. Overall, these results confirm that distilled datasets retain the essential decision structure of the original training distribution — especially the separation between default and non-default classes—whereas random subsampling fails to preserve such structure under moderate to severe class imbalance.

| Dataset | Original AUC | Distilled AUC | Random AUC | ΔAUC | ΔKS | ΔF1-Score | ΔRecall | ΔPrecision |
|---|---|---|---|---|---|---|---|---|
| Bank Personal Loan | 0.969 ±0.034 | 0.924 ±0.098 | 0.649 ±0.135 | 0.276*** | 0.4646*** | 0.5145*** | 0.6348*** | 0.4058*** |
| CHN Banks | 0.808 ±0.044 | 0.738 ±0.103 | 0.62 ±0.168 | 0.118*** | 0.1057*** | 0.1468*** | 0.257*** | 0.0957*** |
| GMC | 0.729 ±0.080 | 0.631 ±0.064 | 0.588 ±0.092 | 0.043*** | 0.0542*** | 0.0843*** | 0.3662*** | -0.0952* |
| Credit Card | 0.746 ±0.031 | 0.663 ±0.083 | 0.553 ±0.102 | 0.110*** | 0.1076*** | 0.1631*** | 0.1584*** | 0.1008*** |
| German Credit | 0.736 ±0.027 | 0.562 ±0.128 | 0.511 ±0.106 | 0.051*** | 0.0739*** | 0.2433*** | 0.373*** | 0.1731*** |
| Lending Club | 0.675 ±0.021 | 0.619 ±0.056 | 0.528 ±0.04 | 0.092*** | 0.1254*** | 0.3233*** | 0.3669*** | 0.2374*** |

**Table 2** Dataset-wise performance of classifiers trained on MSE-distilled versus randomly selected subsets. For all datasets, the distilled data significantly outperform random sampling, with ΔAUC values between +0.04 and +0.28 (all $p < 10^{-3}$, most $p < 10^{-15}$). The baseline is the random selected datasets.

### 4.3.2 Effect of Support Size on Imbalance-Aware Distillation

To understand how the distillation interacts with different imbalance-aware objectives, we compute, for every configuration <dataset, classifier, support size>, the per-configuration AUC improvement

$$\Delta AUC = AUC_{algo} - AUC_{MSE} \tag{8}$$

where algo is the other strategies CE, re-Bn, focal and LA. We assess its significance using paired t-tests across the 5-fold. **Figure 8** summarizes the resulting $\Delta AUC$ values in an objective – support-size matrix, where each cell shows the average Δ AUC over all applicable configurations, accompanied by significance stars indicating directional significance at the 10%, 5%, and 1% levels.

The top subplot -All classifiers, reveals a clear hierarchy among objectives. The focal strategy dominates across almost the entire support range: its Δ AUC is positive for every budget, typically around 0.03 – 0.09 for medium and large support sizes, and almost all units carry ** or *** marks, indicating that improvements over MSE are not only sizable but also highly consistent across datasets and classifiers. LA exhibits a milder yet robust positive effect: Δ AUC is close to zero for very small support sizes, increases to a few percentage points once the distilled set exceeds roughly 100 – 150 samples, and remains weakly positive thereafter.

In contrast, re-Bn and CE rarely produce systematic gains. re-Bn fluctuates around zero with small negative values in many cells, while CE is uniformly detrimental, with Δ AUC often below −0.05 and frequent significance stars in the negative direction. Simply replacing MSE with standard CE is therefore insufficient.

The lower panels show that these patterns are remarkably stable across classifiers. For TabPFN alone, both Focal and LA yield clear gains at almost all support sizes, suggesting that imbalance-aware weighting is complementary to the pretrained tabular prior. For non-TabPFN models, Focal remains strongly positive, whereas the advantage of LA is more modest, consistent with LA primarily correcting prior-probability bias in high-capacity models. Comparing deep models (Neural Network + TabPFN) with tree ensembles (LightGBM + Random Forest) further shows that tree-based classifiers benefit particularly strongly from Focal: Δ AUC is positive and significant across essentially all budgets, reflecting that concentrating distilled samples on hard and minority cases helps trees carve out more informative partitions. CE, on the other hand, is especially harmful for tree models, where almost all units exhibit significantly negative Δ AUC.

#### 4.3.3 Comparison with existing technique (baseline model)

Across all credit datasets and classifiers, the proposed imbalance-aware distillation objectives consistently outperform the standard MSE objective as well as naive random subsampling. In particular, Focal-loss-based distillation achieves higher AUC on the majority of dataset – classifier pairs, with statistically significant gains in many highly imbalanced settings. Logit-adjusted (LA) distillation is especially effective at very small support sizes when combined with TabPFN, while tree-based models trained on Focal-distilled data tend to dominate performance in most other cases. Relative to training classifiers directly on small random subsets of the original data, models trained on our distilled datasets exhibit both higher average performance and lower variance, indicating that the synthetic samples encode more informative, class-imbalanced structure than raw subsamples of the same size.

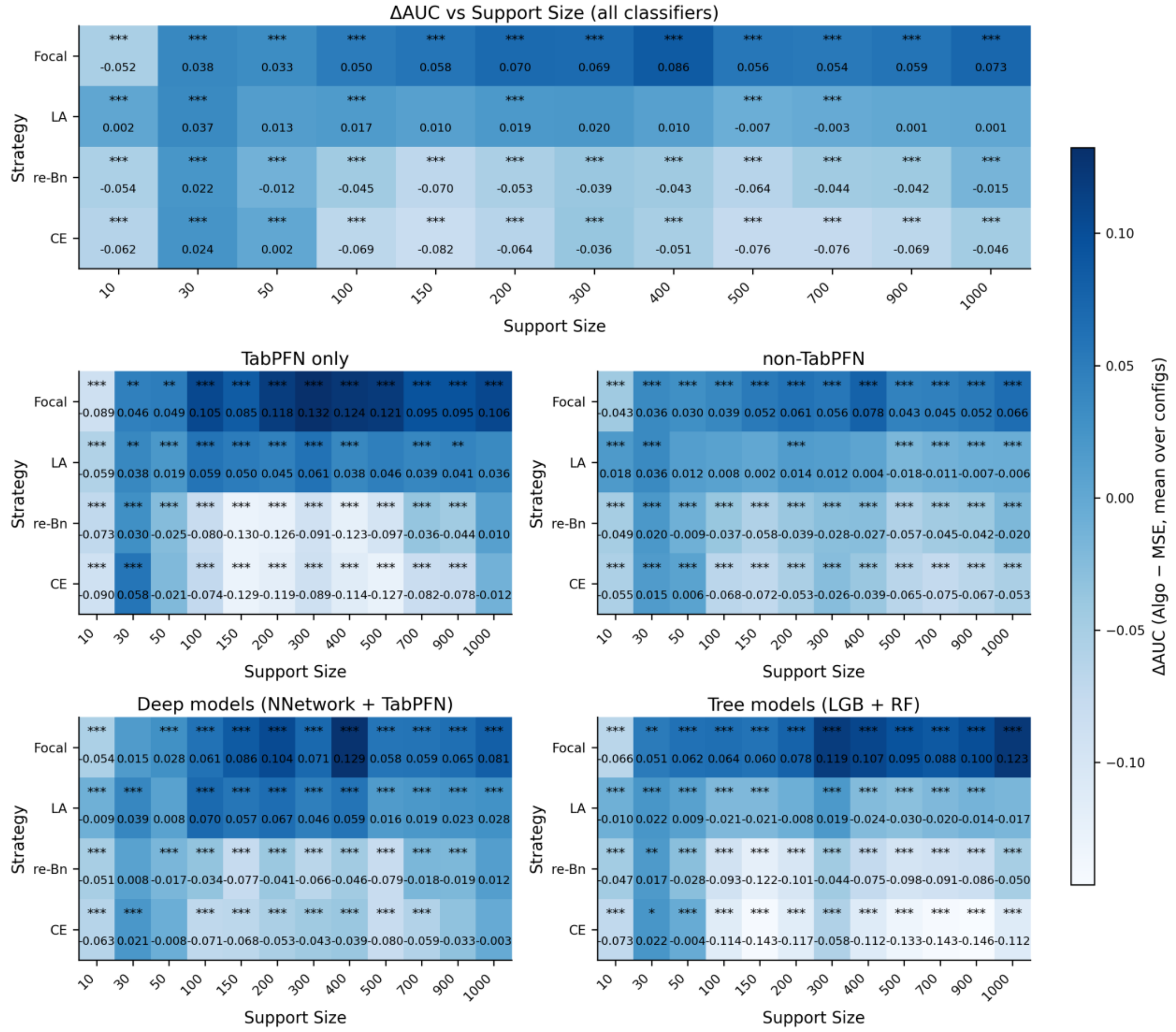

**Figure 8** The △ (strategy − MSE) versus support size across datasets and classifiers. Each heatmap cell reports the average AUC difference between a given distillation strategy (Focal, LA, re-Bn, CE) and the MSE baseline over all ( dataset , classifier ) (dataset,classifier) configurations at that support size, with darker colors indicating larger positive gains. The top panel aggregates over all classifiers; the middle row separates TabPFN-only and non-TabPFN backbones; the bottom row contrasts deep models (Neural Network + TabPFN) with tree ensembles (LightGBM + Random Forest). **Stars denote the directional significance of the mean Δ AUC ΔAUC based on configuration-level paired t-tests across folds (*, , and *** corresponding to the 10%, 5%, and 1% levels, respectively).

Overall, the support-size analyses demonstrate that imbalance-aware distillation objectives do not merely scale with the distillation budget—they fundamentally alter how limited synthetic capacity is allocated. Focal consistently delivers the strongest and most stable improvements; LA offers moderate but reliable gains; and CE-based objectives either neutral or harmful. These patterns persist across datasets, classifiers, and support sizes, underscoring the importance of explicitly handling imbalance within dataset distillation process.

#### 4.3.4 Meta-regression analysis

The heatmaps in **Figure 8** show that imbalance-aware strategies, especially Focal and LA, tend to outperform MSE, but they do not explain why the magnitude of Δ AUC varies across datasets and models. To address this, we treat each configuration as one observation and regress its average Δ AUC on dataset- and model-level covariates (Table 3). The regressors include log class-imbalance (log_imbalance), dataset size (log_samples), feature dimensionality (NumFeatures, label is excluded), classifier indicators (is_TabPFN, is_Tree), and the log support sizes (log_support). We then fit separate OLS meta-regressions for LA, Focal, re-Bn, and CE, using HC3 robust standard errors.

Focal strategy exhibits the clearest and most structured pattern. Its coefficient on log_imbalance is large and positive (0.0750***), indicating that Focal's advantage over MSE grows with the severity of class imbalance. moving from mildly imbalanced datasets to highly imbalanced ones can easily translate into an additional 3 – 7 percentage points of Δ AUC. At the same time, the negative coefficient on log_samples (−0.0677***) shows that these gains are strongest on smaller datasets, where sampling noise is more pronounced. Focal also benefits all major classifiers—both is_TabPFN and is_Tree are strongly positive (0.0595*** and 0.0614***), and log_support is positive (0.0142***), consistent with **Figure 8** where Focal keeps improving as the distilled support size increases. The relatively high adjusted $R^2$ (0.511) indicates that these interpretable covariates account for a substantial portion of the variation in Focal's Δ AUC.

| | **(1) LA** | **(2) Focal** | **(3) re-Bn** | **(4) CE** |
|---|---|---|---|---|
| **log_imbalance** | 0.0057 | 0.0750*** | -0.0386* | -0.1367*** |
| | (0.55) | (5.90) | (-1.95) | (-6.96) |
| **log_samples** | -0.0171*** | -0.0677*** | -0.0202*** | -0.0108* |
| | (-3.15) | (-11.32) | (-3.09) | (-1.72) |
| **NumFeatures** | -0.0006* | 0.0000 | 0.0005 | -0.0022*** |
| | (-1.93) | (0.00) | (0.84) | (-4.06) |
| **is_TabPFN** | 0.0350*** | 0.0595*** | -0.0569*** | -0.0774*** |
| | (3.28) | (4.59) | (-4.70) | (-4.89) |
| **is_Tree** | 0.0053 | 0.0614*** | -0.0496*** | -0.0963*** |
| | (0.59) | (6.95) | (-5.32) | (-9.37) |
| **log_support** | -0.0022 | 0.0142*** | -0.0071** | -0.0165*** |
| | (-0.74) | (4.31) | (-2.53) | (-3.99) |
| **Adj R-squared** | 0.117 | 0.511 | 0.302 | 0.473 |
| **N** | 396 | 396 | 396 | 396 |

**Table 3.** Meta regression of ΔAUC (algorithm − MSE) on dataset, model, and support-set characteristics. The *, **, and *** denote significance at the 10%, 5% and 1% levels, respectively. Constants are not reported. Intercepts are omitted.

LA displays a more modest but still meaningful profile. Its gains are driven primarily by dataset scale: the coefficient on log_samples is significantly negative (−0.0171***), implying that LA is particularly helpful on smaller datasets, while log_imbalance is small and insignificant. Importantly, is_TabPFN is positive and highly significant (0.0350***), whereas is_Tree is near zero and not significant, indicating that LA mainly amplifies the benefit of the pretrained TabPFN classifier rather than tree ensembles. The coefficient on log_support is not significant, which aligns with the heatmaps where LA's advantage plateaus once the distilled support size becomes moderate.

By contrast, re-Bn and CE exhibit a predominantly negative pattern in our meta-regression results. For re-Bn, log_imbalance is negative (−0.0386*), implying that re-Bn becomes increasingly harmful when imbalance is severe. Both classifier indicators are significantly negative (−0.0569*** for is_TabPFN and

−0.0496*** for is_Tree), suggesting a tendency for re-Bn to reduce AUC relative to MSE on more imbalanced datasets and for both TabPFN and tree models within our setting. The negative coefficient on log_support (−0.0071**) further suggests that this degradation may accumulate as the synthetic sample size increases. CE shows an even stronger negative association: its coefficients on log_imbalance, NumFeatures, is_TabPFN is_Tree, and log_support are all negative and highly significant (e.g., −0.1367*** on log_imbalance and −0.0963*** on is_Tree), indicating that, in the empirical regime we study, CE is most often associated with poorer performance on highly imbalanced, high-dimensional datasets and for tree-based or TabPFN classifiers. The relatively high adjusted $\boldsymbol{R}^2$(0.473) for CE underscores that these associations are statistically consistent in our experiments, although they should not be over-interpreted as universally detrimental outside the present experimental design.

Taken together, the meta-regression reinforces and sharpens the conclusions from the heatmaps. Focal is the only strategy whose incremental benefit over MSE increases with imbalance severity, is strongest on smaller datasets, and remains positive across all major backbone families and support sizes. LA offers a milder but robust gain, concentrated in low-sample regimes and primarily for TabPFN. In contrast, re-Bn and especially CE exhibit consistent negative coefficients along key axes of this experimental environment, suggesting that they are poorly matched to the demands of tabular credit-risk distillation. These results reinforce that explicit and properly targeted imbalance modeling is crucial, and that not all imbalance-aware objectives are equally effective for dataset distillation in financial risk modeling.

## 5. Conclusion

In this study, we propose an imbalance-aware dataset distillation framework that injects class-sensitive objectives directly into the distillation process, providing a unified solution to three challenges that frequently co-occur in credit scoring: restrictive data sharing under privacy constraints, the limited scalability of pretrained tabular models, and severe class imbalance.

Firstly, our tabular-tailored distillation procedure offers a practical mechanism for privacy-conscious data exchange: NNDR analysis indicates that the distilled synthetic datasets retain task-relevant structure while remaining geometrically detached from individual customer records, although we do not provide formal differential privacy guarantees. Secondly, distillation serves as an effective compressor for TabPFN and related large tabular models, substantially improving their scalability by replacing large raw datasets with compact synthetic support samples. Thirdly, by incorporating imbalance-aware objectives during distillation, we reshape the synthetic supports to better preserve minority-class geometry. Empirically, imbalance-aware distilled datasets consistently improve AUC over MSE-based distilled datasets and random subsets, with Focal and LA producing the most informative representations under highly skewed credit distributions.

Our results therefore indicate that imbalance-aware dataset distillation can substantially reduce data volume while preserving credit-risk discrimination, making it a promising building block for scalable and privacy-conscious credit scoring pipelines. At the same time, several limitations remain. Our evaluation is restricted to binary default prediction on public benchmark datasets, which are smaller and less heterogeneous than proprietary institutional portfolios; multi-class credit outcomes, temporal drift, and fairness-related considerations are not yet addressed. Future work will extend the framework to larger private datasets via institutional collaborations, explore federated and differentially private distillation for stronger formal guarantees, and integrate calibration- and cost-sensitive objectives to better align distilled models with practical risk-management requirements.

Overall, our findings show that embedding imbalance-aware objectives into the distillation stage systematically outperforms both standard MSE-based distillation and conventional training on small random subsets. In particular, Focal- and LA-based distillation yield robust gains under severe class imbalance, suggesting that carefully designed class-aware synthetic datasets can serve as a competitive alternative to random undersampling and standard cost-sensitive training on raw data in real-world credit-risk modeling.

# 6. References

Alamri, M., & Ykhlef, M. (2024). Hybrid undersampling and oversampling for handling imbalanced credit card data. *IEEE Access, 12*, 14050–14060.

Ali, A. A., Khedr, A. M., El-Bannany, M., & Kanakkayil, S. (2023). A powerful predicting model for financial statement fraud based on optimized XGBoost ensemble learning technique. *Applied Sciences, 13*(4), 2272.

Almubark, I. (2024). Advanced Credit Card Fraud Detection: An Ensemble Learning Using Random Under Sampling and Two-Stage Thresholding. *IEEE Access*.

Andreeva, G., Calabrese, R., & Osmetti, S. A. (2016). A comparative analysis of the UK and Italian small businesses using Generalised Extreme Value models. *European Journal of Operational Research, 249*(2), 506–516.

Bellotti, T., & Crook, J. (2009). Credit scoring with macroeconomic variables using survival analysis. *Journal of the Operational Research Society, 60*(12), 1699–1707.

Cao, K., Wei, C., Gaidon, A., Arechiga, N., & Ma, T. (2019). Learning imbalanced datasets with label-distribution-aware margin loss. *Advances in neural information processing systems, 32*.

Cao, Y., & Zhai, J. (2023). Bridging the gap–the impact of ChatGPT on financial research. *Journal of Chinese Economic and Business Studies, 21*(2), 177–191.

Chen, T., & Guestrin, C. (2016). *Xgboost: A scalable tree boosting system.* Paper presented at the Proceedings of the 22nd acm sigkdd international conference on knowledge discovery and data mining.

Cheng, X., Liu, S., Sun, X., Wang, Z., Zhou, H., Shao, Y., & Shen, H. (2021). Combating emerging financial risks in the big data era: A perspective review. *Fundamental Research, 1*(5), 595–606.

Dastile, X., Celik, T., & Potsane, M. (2020). Statistical and machine learning models in credit scoring: A systematic literature survey. *Applied Soft Computing, 91*, 106263.

Demajo, L. M., Vella, V., & Dingli, A. (2020). Explainable AI for Interpretable Credit Scoring. arXiv:2012.03749. doi:10.48550/arXiv.2012.03749

Devi, D., Biswas, S. K., & Purkayastha, B. (2020). *A review on solution to class imbalance problem: Undersampling approaches.* Paper presented at the 2020 international conference on computational performance evaluation (ComPE).

Devlin, J., Chang, M.-W., Lee, K., & Toutanova, K. (2018). Bert: Pre-training of deep bidirectional transformers for language understanding. *arXiv preprint arXiv:1810.04805*.

Dumitrescu, E., Hué, S., Hurlin, C., & Tokpavi, S. (2022). Machine learning for credit scoring: Improving logistic regression with non-linear decision-tree effects. *European Journal of Operational Research, 297*(3), 1178–1192.

Elhassan, T., & Aljurf, M. (2016). Classification of imbalance data using tomek link (t-link) combined with random under-sampling (rus) as a data reduction method. *Global J Technol Optim S, 1*, 2016.

Fernández, A., Garcia, S., Herrera, F., & Chawla, N. V. (2018). SMOTE for learning from imbalanced data: progress and challenges, marking the 15-year anniversary. *Journal of artificial intelligence research, 61*, 863–905.

Feuer, B., Hegde, C., & Cohen, N. (2023). Scaling tabpfn: Sketching and feature selection for tabular prior-data fitted networks. *arXiv preprint arXiv:2311.10609*.

Gong, X., Sharma, A., Karanam, S., Wu, Z., Chen, T., Doermann, D., & Innanje, A. (2022). *Preserving privacy in federated learning with ensemble cross-domain knowledge distillation.* Paper presented at the Proceedings of the AAAI Conference on Artificial Intelligence.

Gosain, A., & Sardana, S. (2017). *Handling class imbalance problem using oversampling techniques: A review.* Paper presented at the 2017 international conference on advances in computing, communications and informatics (ICACCI).

He, H., Bai, Y., Garcia, E. A., & Li, S. (2008). *ADASYN: Adaptive synthetic sampling approach for imbalanced learning.* Paper presented at the 2008 IEEE international joint conference on neural networks (IEEE world congress on computational intelligence).

He, H., & Garcia, E. A. (2009). Learning from imbalanced data. *IEEE Transactions on knowledge and data engineering, 21*(9), 1263–1284.

Hiew, J. Z. G., Huang, X., Mou, H., Li, D., Wu, Q., & Xu, Y. (2019). BERT-based financial sentiment index and LSTM-based stock return predictability. *arXiv preprint arXiv:1906.09024*.

Hollmann, N., Müller, S., Eggensperger, K., & Hutter, F. (2022). Tabpfn: A transformer that solves small tabular classification problems in a second. *arXiv preprint arXiv:2207.01848*.

Huang, A., Qiu, L., & Li, Z. (2021). Applying deep learning method in TVP-VAR model under systematic financial risk monitoring and early warning. *Journal of Computational and Applied Mathematics, 382*, 113065.

Huang, X., Zhang, C.-Z., & Yuan, J. (2020). Predicting extreme financial risks on imbalanced dataset: A combined kernel FCM and kernel SMOTE based SVM classifier. *Computational Economics, 56*(1), 187–216.

Jing, R., Tian, H., Zhou, G., Zhang, X., Zheng, X., & Zeng, D. D. (2021). *A GNN-based Few-shot learning model on the Credit Card Fraud detection.* Paper presented at the 2021 IEEE 1st International Conference on Digital Twins and Parallel Intelligence (DTPI).

Kang, I., Ram, P., Zhou, Y., Samulowitz, H., & Seneviratne, O. (2024). *Effective data distillation for tabular datasets (Student abstract).* Paper presented at the Proceedings of the AAAI Conference on Artificial Intelligence.

Ke, G., Meng, Q., Finley, T., Wang, T., Chen, W., Ma, W., . . . Liu, T.-Y. (2017). *LightGBM: a highly efficient gradient boosting decision tree*. Paper presented at the Proceedings of the 31st International Conference on Neural Information Processing Systems, Long Beach, California, USA.

Kim, D., Wang, K., Sclaroff, S., & Saenko, K. (2022). *A broad study of pre-training for domain generalization and adaptation.* Paper presented at the European conference on computer vision.

Kute, D. V., Pradhan, B., Shukla, N., & Alamri, A. (2021). Deep learning and explainable artificial intelligence techniques applied for detecting money laundering–a critical review. *IEEE Access, 9*, 82300–82317.

Lei, S., & Tao, D. (2023). A comprehensive survey of dataset distillation. *IEEE Transactions on Pattern Analysis and Machine Intelligence*.

Li, H., & Wu, W. (2024). Loan default predictability with explainable machine learning. *Finance Research Letters, 60*, 104867.

Li, X., Zheng, H., Tao, K., & Mao, M. (2025). Implementation of an Asymmetric Adjusted Activation Function for Class Imbalance Credit Scoring. *arXiv preprint arXiv:2501.12285*.

Li, Y., & Pan, Y. (2022). A novel ensemble deep learning model for stock prediction based on stock prices and news. *International Journal of Data Science and Analytics, 13*(2), 139–149.

Lin, T.-Y., Goyal, P., Girshick, R., He, K., & Dollár, P. (2017). Focal Loss for Dense Object Detection. arXiv:1708.02002. doi:10.48550/arXiv.1708.02002

Liu, H., Mao, M., Li, X., & Gao, J. (2025). Model interpretability on private-safe oriented student dropout prediction. *Plos one, 20*(3), e0317726.

Liu, S.-Y., & Ye, H.-J. (2025). Tabpfn unleashed: A scalable and effective solution to tabular classification problems. *arXiv preprint arXiv:2502.02527*.

Liu, W., Fan, H., Xia, M., & Xia, M. (2022). A focal-aware cost-sensitive boosted tree for imbalanced credit scoring. *Expert Systems with Applications, 208*, 118158.

Liu, Y., Ott, M., Goyal, N., Du, J., Joshi, M., Chen, D., . . . Stoyanov, V. (2019). Roberta: A robustly optimized bert pretraining approach. *arXiv preprint arXiv:1907.11692*.

Lu, Y., Shen, M., Wang, H., Wang, X., van Rechem, C., Fu, T., & Wei, W. (2023). Machine learning for synthetic data generation: a review. *arXiv preprint arXiv:2302.04062*.

Magadán, L., Roldán-Gómez, J., Granda, J., & Suárez, F. (2023). Early fault classification in rotating machinery with limited data using TabPFN. *IEEE Sensors Journal*.

Moscato, V., Picariello, A., & Sperlí, G. (2021). A benchmark of machine learning approaches for credit score prediction. *Expert Systems with Applications, 165*, 113986.

Mushava, J., & Murray, M. (2024). Flexible loss functions for binary classification in gradient-boosted decision trees: An application to credit scoring. *Expert Systems with Applications, 238*, 121876.

Nazareth, N., & Reddy, Y. V. R. (2023). Financial applications of machine learning: A literature review. *Expert Systems with Applications, 219*, 119640.

Nguyen, T., Chen, Z., & Lee, J. (2020). Dataset meta-learning from kernel ridge-regression. *arXiv preprint arXiv:2011.00050*.

Pan, K., Gong, M., Feng, K., & Li, H. (2024). Preserving Privacy in Fine-grained Data Distillation with Sparse Answers for Efficient Edge Computing. *IEEE Internet of Things Journal*.

Qian, H., Ma, P., Gao, S., & Song, Y. (2023). Soft reordering one-dimensional convolutional neural network for credit scoring. *Knowl. Based Syst., 266*, 110414. doi:10.1016/j.knosys.2023.110414

Ruiz-Villafranca, S., Roldán-Gómez, J., Gómez, J. M. C., Carrillo-Mondéjar, J., & Martinez, J. L. (2024). A TabPFN-based intrusion detection system for the industrial internet of things. *The Journal of Supercomputing, 80*(14), 20080–20117.

Shwartz-Ziv, R., & Armon, A. (2022). Tabular data: Deep learning is not all you need. *Information Fusion, 81*, 84–90.

Teng, H., Wang, C., Yang, Q., Chen, X., & Li, R. (2023). Leveraging adversarial augmentation on imbalance data for online trading fraud detection. *IEEE Transactions on Computational Social Systems, 11*(2), 1602–1614.

Thomas, L., Crook, J., & Edelman, D. (2017). *Credit scoring and its applications*: SIAM.

Trivedi, S. K. (2020). A study on credit scoring modeling with different feature selection and machine learning approaches. *Technology in Society, 63*, 101413.

Wang, C., Deng, C., & Wang, S. (2020). Imbalance-XGBoost: leveraging weighted and focal losses for binary label-imbalanced classification with XGBoost. *Pattern Recognition Letters, 136*, 190–197.

Wang, Y., Zhang, Y., Lu, Y., & Yu, X. (2020). A Comparative Assessment of Credit Risk Model Based on Machine Learning——a case study of bank loan data. *Procedia Computer Science, 174*, 141–149.

Wu, S., Irsoy, O., Lu, S., Dabravolski, V., Dredze, M., Gehrmann, S., Mann, G. (2023). Bloomberggpt: A large language model for finance. *arXiv preprint arXiv:2303.17564*.

Xu, D., Cirit, O., Asadi, R., Sun, Y., & Wang, W. (2024). Mixture of In-Context Prompters for Tabular PFNs. *arXiv preprint arXiv:2405.16156*.

Yang, H., Liu, X.-Y., & Wang, C. D. (2023). Fingpt: Open-source financial large language models. *arXiv preprint arXiv:2306.06031*.

Yu, R., Liu, S., & Wang, X. (2023). Dataset distillation: A comprehensive review. *IEEE Transactions on pattern analysis and machine intelligence, 46*(1), 150–170.

Zhang, L., & Song, Q. (2022). Multimodel integrated enterprise credit evaluation method based on attention mechanism. *Computational Intelligence and Neuroscience, 2022*(1), 8612759.

Zhao, B., Mopuri, K. R., & Bilen, H. (2020). Dataset condensation with gradient matching. *arXiv preprint arXiv:2006.05929*.

Zhao, H., Liu, Z., Wu, Z., Li, Y., Yang, T., Shu, P., . . . Mai, G. (2024). Revolutionizing finance with llms: An overview of applications and insights. *arXiv preprint arXiv:2401.11641*.

Zhao, L., Li, L., Zheng, X., & Zhang, J. (2021). *A BERT based sentiment analysis and key entity detection approach for online financial texts.* Paper presented at the 2021 IEEE 24th International conference on computer supported cooperative work in design (CSCWD).

Zhou, L., Fujita, H., Ding, H., & Ma, R. (2021). Credit risk modeling on data with two timestamps in peer-to-peer lending by gradient boosting. *Applied Soft Computing, 110*, 107672.

# Supporting Materials

## 1.Statistical descriptions of datasets used in this study

| Feature | Dtype | VarType | Num Unique | Min | Max | Mean | Std |
|---|---|---|---|---|---|---|---|
| **Bank_Personal_Loan** | | | | | | | |
| Age | int | continuous | 45 | 23 | 67 | 45.2 | 11.5 |
| Experience | int | continuous | 47 | -3 | 43 | 20 | 11.5 |
| Income | int | continuous | 162 | 8 | 224 | 73.3 | 45.8 |
| ZIP Code | int | continuous | 453 | 9,307 | 96,651 | 93,134 | 2,337 |
| Family | int | discrete | 4 | 1 | 4 | 2.4 | 1.2 |
| CCAvg | float | continuous | 104 | - | 10 | 1.9 | 1.7 |
| Education | int | discrete | 3 | 1 | 3 | 1.9 | 0.8 |
| Mortgage | int | continuous | 295 | - | 617 | 55.8 | 100.2 |
| Securities Account | int | discrete | 2 | - | 1 | 0.1 | 0.3 |
| CD Account | int | discrete | 2 | - | 1 | 0.1 | 0.2 |
| Online | int | discrete | 2 | - | 1 | 0.6 | 0.5 |
| CreditCard | int | discrete | 2 | - | 1 | 0.3 | 0.5 |
| label | int | discrete | 2 | - | 1 | 0.1 | 0.3 |
| **credit_card** | | | | | | | |
| LIMIT_BAL | int | continuous | 64 | 10,000 | 700,000 | 167,318 | 130,245 |
| SEX | int | discrete | 2 | 1 | 2 | 1.6 | 0.5 |
| EDUCATION | int | discrete | 7 | - | 6 | 1.8 | 0.8 |
| MARRIAGE | int | discrete | 4 | - | 3 | 1.6 | 0.5 |
| AGE | int | continuous | 49 | 21 | 72 | 35.4 | 9.1 |
| PAY_0 | int | discrete | 11 | -2 | 8 | 0 | 1.1 |
| PAY_2 | int | discrete | 10 | -2 | 7 | -0.1 | 1.2 |
| PAY_3 | int | discrete | 9 | -2 | 7 | -0.2 | 1.2 |
| PAY_4 | int | discrete | 8 | -2 | 7 | -0.2 | 1.2 |
| PAY_5 | int | discrete | 8 | -2 | 7 | -0.3 | 1.2 |
| PAY_6 | int | discrete | 8 | -2 | 7 | -0.3 | 1.2 |
| BILL_AMT1 | int | continuous | 2,640 | -154,973 | 621,749 | 49,464 | 73,404 |
| BILL_AMT2 | int | continuous | 2,629 | -4,341 | 624,475 | 47,590 | 69,562 |
| BILL_AMT3 | int | continuous | 2,584 | -15,641 | 632,041 | 45,810 | 66,974 |
| BILL_AMT4 | int | continuous | 2,529 | -170,000 | 516,575 | 42,493 | 62,880 |
| BILL_AMT5 | int | continuous | 2,515 | -61,372 | 493,062 | 39,383 | 58,635 |
| BILL_AMT6 | int | continuous | 2,458 | -16,370 | 474,301 | 38,035 | 57,460 |
| PAY_AMT1 | int | continuous | 1,441 | - | 225,066 | 5,277 | 13,254 |
| PAY_AMT2 | int | continuous | 1,466 | - | 358,689 | 6,023 | 18,539 |
| PAY_AMT3 | int | continuous | 1,336 | - | 344,261 | 5,206 | 15,553 |
| PAY_AMT4 | int | continuous | 1,266 | - | 497,000 | 5,009 | 19,182 |

| Feature | Dtype | VarType | Num Unique | Min | Max | Mean | Std |
|---|---|---|---|---|---|---|---|
| PAY_AMT5 | int | continuous | 1,228 | - | 195,829 | 4,537 | 13,198 |
| PAY_AMT6 | int | continuous | 1,243 | - | 351,282 | 4,740 | 15,567 |
| label | int | discrete | 2 | - | 1 | 0.2 | 0.4 |
| **Lending Club** | | | | | | | |
| label | int | discrete | 2 | - | 1 | 0.4 | 0.5 |
| loan_amnt | int | continuous | 598 | 1,000 | 37,500 | 15,111 | 8,451 |
| term | int | discrete | 2 | 36 | 60 | 43.9 | 11.3 |
| int_rate | float | continuous | 218 | 5.3 | 30.8 | 15.1 | 4.6 |
| installment | float | continuous | 2,441 | 32.4 | 1,327.50 | 456.8 | 248.3 |
| annual_inc | float | continuous | 587 | 9,600 | 250,000 | 68,856 | 35,519 |
| dti | float | continuous | 1,938 | - | 49.4 | 19.3 | 8.4 |
| earliest_cr_line | int | continuous | 46 | 1,965 | 2,012 | 1,998 | 7.4 |
| open_acc | int | continuous | 36 | 1 | 37 | 11.6 | 5 |
| pub_rec | int | discrete | 2 | - | 1 | 0.2 | 0.4 |
| revol_bal | int | continuous | 2,831 | - | 187,745 | 16,026 | 15,542 |
| revol_util | float | continuous | 895 | - | 101.7 | 57.2 | 23.2 |
| total_acc | int | continuous | 67 | 3 | 75 | 25.2 | 11.5 |
| mort_acc | int | discrete | 2 | - | 1 | 0.6 | 0.5 |
| pub_rec_bankruptcies | int | discrete | 2 | - | 1 | 0.1 | 0.3 |
| sub_grade_A2 | int | discrete | 2 | - | 1 | 0 | 0.1 |
| sub_grade_A3 | int | discrete | 2 | - | 1 | 0 | 0.1 |
| sub_grade_A4 | int | discrete | 2 | - | 1 | 0 | 0.1 |
| sub_grade_A5 | int | discrete | 2 | - | 1 | 0 | 0.2 |
| sub_grade_B1 | int | discrete | 2 | - | 1 | 0 | 0.2 |
| sub_grade_B2 | int | discrete | 2 | - | 1 | 0 | 0.2 |
| sub_grade_B3 | int | discrete | 2 | - | 1 | 0.1 | 0.2 |
| sub_grade_B4 | int | discrete | 2 | - | 1 | 0 | 0.2 |
| sub_grade_B5 | int | discrete | 2 | - | 1 | 0 | 0.2 |
| sub_grade_C1 | int | discrete | 2 | - | 1 | 0.1 | 0.2 |
| sub_grade_C2 | int | discrete | 2 | - | 1 | 0.1 | 0.2 |
| sub_grade_C3 | int | discrete | 2 | - | 1 | 0.1 | 0.2 |
| sub_grade_C4 | int | discrete | 2 | - | 1 | 0.1 | 0.2 |
| sub_grade_C5 | int | discrete | 2 | - | 1 | 0.1 | 0.2 |
| sub_grade_D1 | int | discrete | 2 | - | 1 | 0 | 0.2 |
| sub_grade_D2 | int | discrete | 2 | - | 1 | 0 | 0.2 |
| sub_grade_D3 | int | discrete | 2 | - | 1 | 0 | 0.2 |
| sub_grade_D4 | int | discrete | 2 | - | 1 | 0 | 0.2 |
| sub_grade_D5 | int | discrete | 2 | - | 1 | 0 | 0.2 |
| sub_grade_E1 | int | discrete | 2 | - | 1 | 0 | 0.2 |
| sub_grade_E2 | int | discrete | 2 | - | 1 | 0 | 0.2 |

| Feature | Dtype | VarType | Num Unique | Min | Max | Mean | Std |
|---|---|---|---|---|---|---|---|
| sub_grade_E3 | int | discrete | 2 | - | 1 | 0 | 0.2 |
| sub_grade_E4 | int | discrete | 2 | - | 1 | 0 | 0.1 |
| sub_grade_E5 | int | discrete | 2 | - | 1 | 0 | 0.1 |
| sub_grade_F1 | int | discrete | 2 | - | 1 | 0 | 0.1 |
| sub_grade_F2 | int | discrete | 2 | - | 1 | 0 | 0.1 |
| sub_grade_F3 | int | discrete | 2 | - | 1 | 0 | 0.1 |
| sub_grade_F4 | int | discrete | 2 | - | 1 | 0 | 0.1 |
| sub_grade_F5 | int | discrete | 2 | - | 1 | 0 | 0.1 |
| sub_grade_G1 | int | discrete | 2 | - | 1 | 0 | 0.1 |
| sub_grade_G2 | int | discrete | 2 | - | 1 | 0 | 0.1 |
| sub_grade_G3 | int | discrete | 2 | - | 1 | 0 | 0.1 |
| sub_grade_G4 | int | discrete | 2 | - | 1 | 0 | 0 |
| sub_grade_G5 | int | discrete | 2 | - | 1 | 0 | 0 |
| verification_status_Source Verified | int | discrete | 2 | - | 1 | 0.4 | 0.5 |
| verification_status_Verified | int | discrete | 2 | - | 1 | 0.4 | 0.5 |
| purpose_credit_card | int | discrete | 2 | - | 1 | 0.2 | 0.4 |
| purpose_debt_consolidation | int | discrete | 2 | - | 1 | 0.6 | 0.5 |
| purpose_educational | int | discrete | 1 | - | - | - | - |
| **German Credit** | | | | | | | |
| Account_Balance | int | discrete | 4 | 1 | 4 | 2.6 | 1.3 |
| Duration_of_Credit_monthly | int | continuous | 32 | 4 | 72 | 20.1 | 11.7 |
| Payment_Status_of_Previous_Credit | int | discrete | 5 | - | 4 | 2.5 | 1.1 |
| Purpose | int | discrete | 10 | - | 10 | 2.8 | 2.8 |
| Credit_Amount | int | continuous | 609 | 250 | 18,424 | 3,219 | 2,781 |
| Value_Savings_Stocks | int | discrete | 5 | 1 | 5 | 2.1 | 1.6 |
| Length_of_current_employment | int | discrete | 5 | 1 | 5 | 3.4 | 1.2 |
| Instalment_per_cent | int | discrete | 4 | 1 | 4 | 2.9 | 1.1 |
| Sex_Marital_Status | int | discrete | 4 | 1 | 4 | 2.7 | 0.7 |
| Guarantors | int | discrete | 3 | 1 | 3 | 1.1 | 0.5 |

| Feature | Dtype | VarType | Num Unique | Min | Max | Mean | Std |
|---|---|---|---|---|---|---|---|
| Duration_in_Current_address | int | discrete | 4 | 1 | 4 | 2.8 | 1.1 |
| Most_valuable_available_asset | int | discrete | 4 | 1 | 4 | 2.4 | 1.1 |
| Age_years | int | continuous | 53 | 19 | 75 | 35.6 | 11.7 |
| Concurrent_Credits | int | discrete | 3 | 1 | 3 | 2.7 | 0.7 |
| Type_of_apartment | int | discrete | 3 | 1 | 3 | 1.9 | 0.5 |
| No_of_Credits_at_this_Bank | int | discrete | 4 | 1 | 4 | 1.4 | 0.6 |
| Occupation | int | discrete | 4 | 1 | 4 | 2.9 | 0.7 |
| No_of_dependents | int | discrete | 2 | 1 | 2 | 1.2 | 0.4 |
| Telephone | int | discrete | 2 | 1 | 2 | 1.4 | 0.5 |
| Foreign_Worker | int | discrete | 2 | 1 | 2 | 1 | 0.2 |
| label | int | discrete | 2 | - | 1 | 0.3 | 0.5 |
| **CHN_Banks** | | | | | | | |
| 1 | float | continuous | 73 | - | 1 | 0.4 | 0.2 |
| 2 | float | continuous | 2,367 | - | 0.7 | 0 | 0 |
| 3 | float | continuous | 363 | - | 0.4 | 0 | 0 |
| 4 | float | discrete | 20 | - | 0.3 | 0 | 0 |
| 5 | float | continuous | 1,457 | - | 0.7 | 0 | 0 |
| 6 | float | continuous | 42 | - | 1 | 0 | 0 |
| 7 | float | continuous | 926 | - | 0.2 | 0 | 0 |
| 8 | float | discrete | 6 | - | 0.2 | 0 | 0 |
| 9 | float | continuous | 29 | - | 0.3 | 0 | 0 |
| 10 | float | continuous | 794 | - | 0.1 | 0 | 0 |
| 11 | float | continuous | 1,190 | - | 0.3 | 0 | 0 |
| 12 | float | continuous | 1,494 | - | 0.7 | 0 | 0 |
| 13 | float | continuous | 1,408 | - | 0.5 | 0 | 0 |
| 14 | float | continuous | 1,318 | - | 0.3 | 0 | 0 |
| 15 | float | continuous | 1,265 | - | 0.4 | 0 | 0 |
| 16 | float | continuous | 801 | - | 0.5 | 0 | 0 |
| 17 | float | continuous | 926 | - | 0.2 | 0 | 0 |
| 18 | float | continuous | 1,024 | - | 0.3 | 0 | 0 |
| 19 | float | continuous | 1,133 | - | 0.3 | 0 | 0 |
| 20 | float | continuous | 84 | - | 0.4 | 0 | 0 |
| 21 | float | continuous | 466 | - | 0.2 | 0 | 0 |
| 22 | float | continuous | 1,137 | - | 0.7 | 0 | 0 |
| 23 | float | discrete | 4 | - | 1 | 0 | 0 |
| 24 | float | discrete | 4 | - | 1 | 0 | 0 |

| Feature | Dtype | VarType | Num Unique | Min | Max | Mean | Std |
|---|---|---|---|---|---|---|---|
| 25 | float | discrete | 11 | - | 0.3 | 0 | 0 |
| 26 | float | continuous | 1,242 | - | 0.8 | 0.1 | 0.2 |
| 27 | float | continuous | 361 | - | 1 | 0 | 0 |
| 28 | float | continuous | 2,741 | - | 1 | 0.3 | 0.2 |
| 29 | float | continuous | 2,741 | - | 1 | 0.3 | 0.2 |
| 30 | int | discrete | 2 | - | 1 | 0.8 | 0.4 |
| 31 | float | discrete | 3 | - | 1 | 0.5 | 0.5 |
| 32 | int | discrete | 2 | - | 1 | 0.4 | 0.5 |
| 33 | int | discrete | 2 | - | 1 | 0 | 0.1 |
| 34 | int | discrete | 2 | - | 1 | 0.1 | 0.3 |
| 35 | int | discrete | 2 | - | 1 | 0 | 0.1 |
| 36 | int | discrete | 2 | - | 1 | 0 | 0 |
| 37 | float | discrete | 7 | - | 1 | 0.6 | 0.2 |
| label | int | discrete | 2 | - | 1 | 0.2 | 0.4 |
| **GMC** | | | | | | | |
| label | int | discrete | 2 | - | 1 | 0.1 | 0.3 |
| RevolvingUtilizationOfUnsecuredLines | float | continuous | 2,617 | - | 1800 | 1 | 36 |
| age | int | continuous | 77 | - | 99 | 51 | 14.6 |
| NumberOfTime30-59DaysPastDueNotWorse | int | discrete | 9 | - | 98 | 0.4 | 3.6 |
| DebtRatio | float | continuous | 2,931 | - | 6,836 | 22 | 259 |
| MonthlyIncome | int | continuous | 1,450 | - | 71,000 | 6,345 | 4,949 |
| NumberOfOpenCreditLinesAndLoans | int | continuous | 38 | - | 57 | 8.8 | 5.4 |
| NumberOfTimes90DaysLate | int | discrete | 10 | - | 98 | 0.2 | 3.6 |
| NumberRealEstateLoansOrLines | int | discrete | 11 | - | 21 | 1 | 1.1 |
| NumberOfTime60-89DaysPastDueNotWorse | int | discrete | 6 | - | 98 | 0.2 | 3.6 |
| NumberOfDependents | int | discrete | 8 | - | 9 | 0.8 | 1.1 |

**Table S1.** Statistical description of various datasets

## 2.NNDR evaluations on all datasets

The NNDR of all dataset distilled from different strategies, along with the SMOTE method and the Random selected samples. For each dataset and support size, the boxplots report the distribution of Nearest-Neighbor Distance Ratios (NNDR) between each synthetic point and its two closest neighbors in the original training set. Values closer to 1 indicate stronger geometric separation from real samples (higher privacy margin), whereas values near 0 imply synthetic points collapsing toward specific training instances.

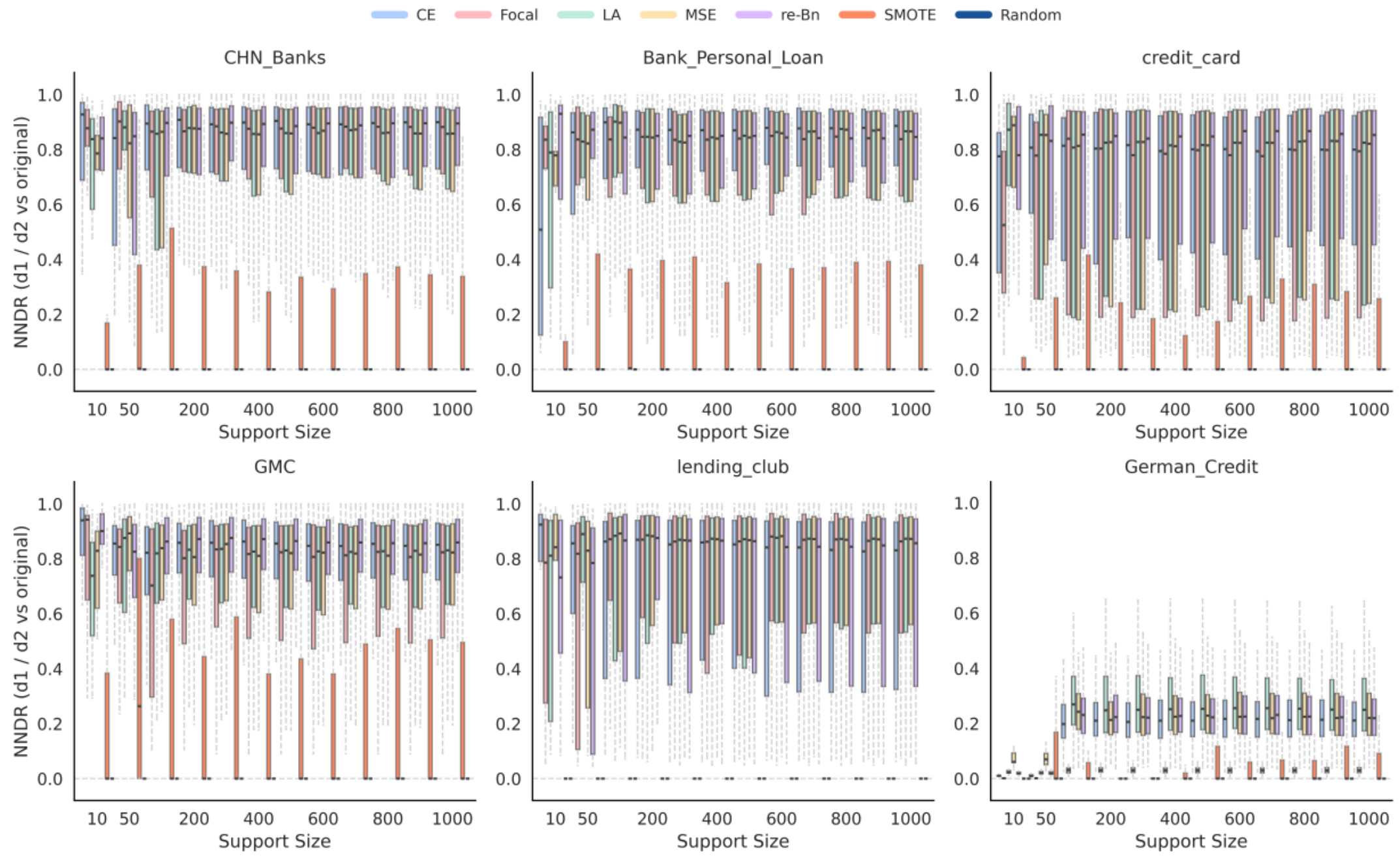

**Figure S1**. NNDR comparison across distillation objectives, datasets, and support sizes.

## 3.Computational cost for all datasets

| Support Size | mean | std |
|---:|---:|---:|
| 10 | 1.842 | 0.540 |
| 50 | 1.743 | 0.139 |
| 100 | 1.826 | 0.253 |
| 150 | 1.900 | 0.163 |
| 200 | 1.934 | 0.184 |
| 300 | 2.027 | 0.244 |
| 400 | 2.195 | 0.280 |
| 500 | 2.345 | 0.405 |
| 600 | 2.445 | 0.440 |
| 700 | 2.564 | 0.473 |
| 800 | 2.666 | 0.440 |
| 900 | 2.798 | 0.577 |
| 1000 | 2.888 | 0.683 |

**Table S2.** Computational cost on each support size. The mean is represented by the average runtime value of the combination of <dataset, classifier, strategy>.

## 4. AUC performance on various datasets

| Dataset | Original AUC | Distilled AUC | Distilled/Original | Sample Ratio | Random AUC | Random/Original |
|---|---|---|---|---|---|---|
| Bank Personal Loan | 0.969 | 0.924 | 95.4% | 31.3% | 0.649 | 67.0% |
| CHN Banks | 0.808 | 0.738 | 91.3% | 1.6% | 0.62 | 76.7% |
| GMC | 0.729 | 0.631 | 86.6% | 0.7% | 0.588 | 80.7% |
| Credit Card | 0.746 | 0.663 | 88.9% | 3.3% | 0.553 | 74.1% |
| German Credit | 0.736 | 0.562 | 76.4% | -- | 0.511 | -- |
| Lending Club | 0.675 | 0.619 | 91.7% | 6.8% | 0.528 | 78.2% |

**Table S3.** Quantitative value of the performance comparison between models trained on full datasets (Original), distilled synthetic datasets (Distilled), and randomly subsampled real data (Random) across six credit-scoring benchmarks. Distilled/Original and Random/Original demonstrate the AUC ratios relative to full-data training, and Sample Ratio denotes the proportion of distilled samples relative to the original training set.

Across six credit-scoring datasets, the distilled synthetic sets preserve a remarkably high fraction of the predictive performance obtained using full data, despite using only a small proportion of the original samples. Distilled AUC ranges from 0.562 to 0.924, corresponding to 76.4%–95.4% of the full-data AUC, while relying on merely 0.7%–31.3% of the original training instances.

In contrast, randomly subsampled real data are consistently weaker, achieving only 67.0%–80.7% of the full-data AUC. For example, on Bank Personal Loan, the distilled set achieves an AUC of 0.924 (95.4% of the original) using 31.3% of the data, whereas random subsampling yields only 0.649 (67.0%). Even on more challenging datasets such as GMC and Credit Card, distilled datasets outperform random subsets by clear margins, indicating that dataset distillation identifies compact synthetic prototypes that retain task-relevant structure far more effectively than naïve downsampling. Due to the original sample size of German Credit is smaller than the support size ( when support size goes high), we resampled the data points with replacement, such that the sample ratio is not representative and we did not demonstrate it.

**5. Model performance metrics**

| **Metrics** | **Datasets** | Bank Personal Loan | | CHN Banks | | GMC | | German Credit | | Credit card | | Lending club | |
|---|---|---|---|---|---|---|---|---|---|---|---|---|---|
| AUC | Original | 0.969 | ±0.034 | 0.808 | ±0.044 | 0.729 | ±0.08 | 0.746 | ±0.031 | 0.736 | ±0.027 | 0.675 | ±0.021 |
| | Distilled | 0.924 | ±0.098 | 0.738 | ±0.103 | 0.631 | ±0.064 | 0.696 | ±0.088 | 0.663 | ±0.083 | 0.619 | ±0.056 |
| | Random | 0.649 | ±0.135 | 0.620 | ±0.168 | 0.588 | ±0.092 | 0.561 | ±0.096 | 0.553 | ±0.102 | 0.527 | ±0.04 |
| KS | Original | 0.872 | ±0.077 | 0.500 | ±0.07 | 0.374 | ±0.118 | 0.409 | ±0.066 | 0.376 | ±0.038 | 0.271 | ±0.035 |
| | Distilled | 0.758 | ±0.187 | 0.391 | ±0.126 | 0.224 | ±0.092 | 0.347 | ±0.126 | 0.277 | ±0.1 | 0.195 | ±0.08 |
| | Random | 0.293 | ±0.22 | 0.285 | ±0.176 | 0.169 | ±0.148 | 0.161 | ±0.143 | 0.170 | ±0.137 | 0.069 | ±0.059 |
| F1-Score | Original | 0.814 | ±0.113 | 0.323 | ±0.038 | 0.153 | ±0.072 | 0.483 | ±0.047 | 0.421 | ±0.033 | 0.461 | ±0.022 |
| | Distilled | 0.625 | ±0.177 | 0.400 | ±0.07 | 0.166 | ±0.03 | 0.533 | ±0.065 | 0.429 | ±0.06 | 0.487 | ±0.128 |
| | Random | 0.111 | ±0.14 | 0.253 | ±0.172 | 0.082 | ±0.092 | 0.323 | ±0.208 | 0.266 | ±0.161 | 0.164 | ±0.212 |
| Recall | Original | 0.741 | ±0.149 | 0.240 | ±0.043 | 0.094 | ±0.051 | 0.400 | ±0.053 | 0.316 | ±0.037 | 0.388 | ±0.031 |
| | Distilled | 0.855 | ±0.129 | 0.768 | ±0.125 | 0.673 | ±0.172 | 0.697 | ±0.092 | 0.583 | ±0.102 | 0.570 | ±0.186 |
| | Random | 0.220 | ±0.361 | 0.511 | ±0.361 | 0.306 | ±0.396 | 0.503 | ±0.355 | 0.424 | ±0.293 | 0.203 | ±0.302 |
| Precision | Original | 0.918 | ±0.058 | 0.525 | ±0.055 | 0.553 | ±0.028 | 0.632 | ±0.051 | 0.644 | ±0.055 | 0.576 | ±0.034 |
| | Distilled | 0.520 | ±0.195 | 0.273 | ±0.056 | 0.097 | ±0.022 | 0.443 | ±0.083 | 0.353 | ±0.079 | 0.452 | ±0.055 |
| | Random | 0.114 | ±0.142 | 0.177 | ±0.123 | 0.192 | ±0.245 | 0.260 | ±0.164 | 0.253 | ±0.211 | 0.215 | ±0.182 |

**Table S4**. The summary statistics of model performance across five evaluation folds for full-data training (Original), imbalance-aware distilled datasets (Distilled), and randomly subsampled real data (Random) on six credit-scoring datasets. Metrics include AUC, KS, F1-score, Recall, and Precision, each reported as mean ± standard deviation. The results show that distilled datasets substantially outperform random subsets while retaining a high proportion of the full-data performance across all datasets and all metrics.